\documentclass[10pt,twocolumn,letterpaper]{article}

\usepackage{iccv}
\usepackage{times}
\usepackage{epsfig}
\usepackage{lipsum}
\usepackage{graphicx}
\usepackage{dblfloatfix}
\usepackage{amsmath}
\usepackage{amssymb}
\usepackage{bm}
\usepackage{mathrsfs}
\usepackage{breqn}
\usepackage{physics}
\usepackage{multirow}
\usepackage{subcaption}

\makeatletter
\DeclareRobustCommand\onedot{\futurelet\@let@token\@onedot}
\def\@onedot{\ifx\@let@token.\else.\null\fi\xspace}
\def\eg{\emph{e.g}\onedot} 
\def\ie{\emph{i.e}\onedot} 
 
\def\etc{\emph{etc}\onedot}

\makeatother

\DeclareMathOperator*{\argmin}{\arg\!\min}
\newcommand\customparagraph[1]{\vspace{0.6em}\noindent\textbf{#1}}

\newcommand{\Tref}[1]{Table~\ref{#1}}
\newcommand{\Eref}[1]{Equation~(\ref{#1})}

\newcommand{\Sref}[1]{Section~\ref{#1}}

\newcommand{\fref}[1]{Fig.~\ref{#1}}

\usepackage[square, comma, sort&compress, numbers]{natbib} 
\usepackage[english]{babel} 

\usepackage[pagebackref=true,breaklinks=true,letterpaper=true,colorlinks,bookmarks=false]{hyperref}

\iccvfinalcopy 

\def\iccvPaperID{****} 

\begin{document}

\title{Event-driven Video Frame Synthesis}

\author{Zihao W. Wang$^1$\quad 
Weixin Jiang$^1$\quad Kuan He$^1$\quad Boxin Shi$^2$\quad Aggelos Katsaggelos$^1$\quad Oliver Cossairt\\
$^1$ Northwestern University\quad $^2$ Peking University\\
{\tt\small \{winswang, weixinjiang2022\}@u.northwestern.edu}
}

\maketitle

\begin{abstract}
Temporal Video Frame Synthesis (TVFS) aims at synthesizing novel frames at timestamps different from existing frames, which has wide applications in video codec, editing and analysis. In this paper, we propose a high framerate TVFS framework which takes hybrid input data from a low-speed frame-based sensor and a high-speed event-based sensor. Compared to frame-based sensors, event-based sensors report brightness changes at very high speed, which may well provide useful spatio-temoral information for high framerate TVFS. In our framework, we first introduce a differentiable forward model to approximate the physical sensing process, fusing the two different modes of data as well as unifying a variety of TVFS tasks, \ie, interpolation, prediction and motion deblur. We leverage autodifferentiation which propagates the gradients of a loss defined on the measured data back to the latent high framerate video. We show results with better performance compared to state-of-the-art. Second, we develop a deep learning-based strategy to enhance the results from the first step, which we refer as a residual ``denoising'' process. Our trained ``denoiser'' is beyond Gaussian denoising and shows properties such as contrast enhancement and motion awareness. We show that our framework is capable of handling challenging scenes including both fast motion and strong occlusions. Supplementary material, demo and code are released at: \href{https://github.com/winswang/int-event-fusion/tree/win10}{https://github.com/winswang/int-event-fusion/tree/win10}. 
\end{abstract}

\section{Introduction}
Conventional video cameras capture intensity signals at fixed speed and output signals frame by frame. However, this capture convention is motion agnostic. When the motion in the scene is significantly faster than the capturing speed, the motion is usually under-sampled, resulting in motion blur or large discrepancies between consecutive frames, depending on the shutter speed (exposure time). One direct solution to capture fast motion is to use high speed cameras, in exchange with increased hardware complexity, degraded spatial resolution and/or reduced signal-to-noise ratio. Moreover, high speed moments usually happen instantaneously between regular motion. As a consequence, either we end up collecting long sequences of frames with a great amount of redundancy, or the high-speed moment is missed before we realize to turn on the ``slow-motion'' mode.

\begin{figure*}
    \centering
    \includegraphics[width=\linewidth]{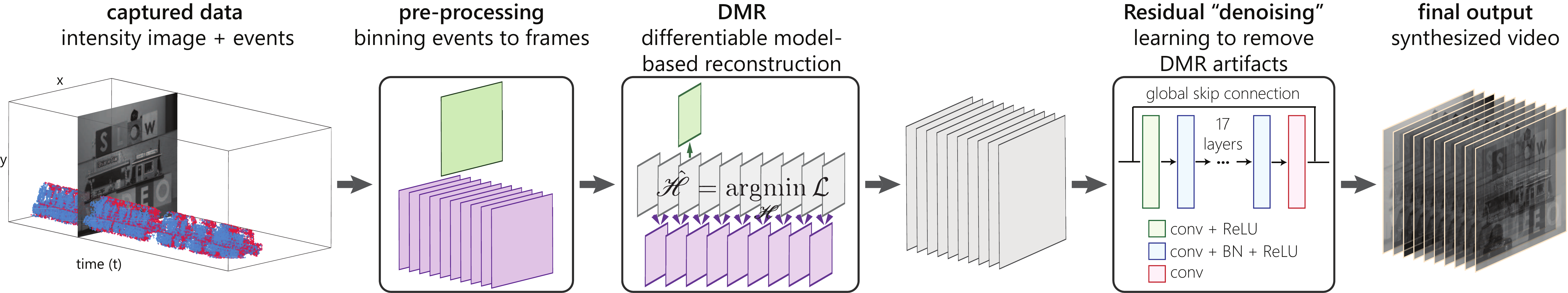}
    \caption{We propose a fusion famework of intensity image(s) and events for high-speed video synthesis. Our synthesis process includes a differentiable model-based reconstruction and a residual ``denoising'' process.}
    \label{fig:overview}
\end{figure*}

We argue that high speed motion can be acquired and synthesized effectively by augmenting a regular-speed camera with a bio-inspired event camera \cite{brandli2014240, lichtsteiner2006128}. Compared to conventional frame-based sensors, event pixels \emph{independently} detect logarithmic brighness variation over time and output ``events'' with four attributes: 2D spatial location, polarity (\eg, ``1'': brightness increases; ``0'': brightness decreases) and timestamp ($\sim1\mu s$ latency). This new sensing modality has salient advantages over frame-based cameras: 1) the asynchronism of event pixels results in sub-millisecond temporal resolution, much higher than regular-speed cameras ($\sim 30$ FPS); 2) since each pixel responds only to intensity changes, the temporal redundancy and power consumption can be significantly reduced; 3) sensing intensity changes in logarithmic scale enlarges dynamic range to over 120 dB\footnote{Typical dynamic range of a conventional camera is 90 dB}. However, event-based cameras have increased noise-level over low framerate cameras. And the bipolar form of output does not represent the exact temporal gradients, introducing challenges for high framerate video reconstruction from event-based cameras alone.

In this paper, we propose a high framerate video synthesis framework using a combination of regular-speed intensity frame(s) and neighboring event streams, as shown in \fref{fig:overview}. Compared to intensity-only or event-only TVFS algorithms, our work takes advantages from both ends, \ie, high-speed information from events and high contrast spatial features from intensity frame(s). 

Our contributions are listed below: 
\begin{enumerate}
\item We introduce a differentiable fusion model enabling to solve various temporal settings. We consider three fundamental cases, \ie, interpolation, prediction and motion deblur, which can serve as building blocks for other complex settings. The problem can be solved by automatic differentiation. We refer to this process as Differentiable Model-based Reconstruction (DMR).
\item We introduce a novel event binning strategy and compare it against conventional stacking-based binning strategy \cite{bardow2016simultaneous, barua2016direct, rebecq2019events, wang2018event}. Our binning preserves the temporal information of events necessary for high frame-rate video reconstruction. Additionally, we perform statistical evaluation for our binning strategy on the existing dataset \cite{mueggler2017event}.
\item We introduce a deep learning strategy for further improving the DMR results. We model the DMR artifacts as additive ``noise'' and perform ``denoising'' via deep residual learning. During training, we augment the samples by randomizing \emph{all} the parameters of the DMR. We show preliminary results that the trained residual denoiser (RD) has properties including constrast enhancement and motion awareness, which is beyond a Gaussian denoiser.
\end{enumerate}


\begin{figure}
    \centering
    \includegraphics[width=\linewidth]{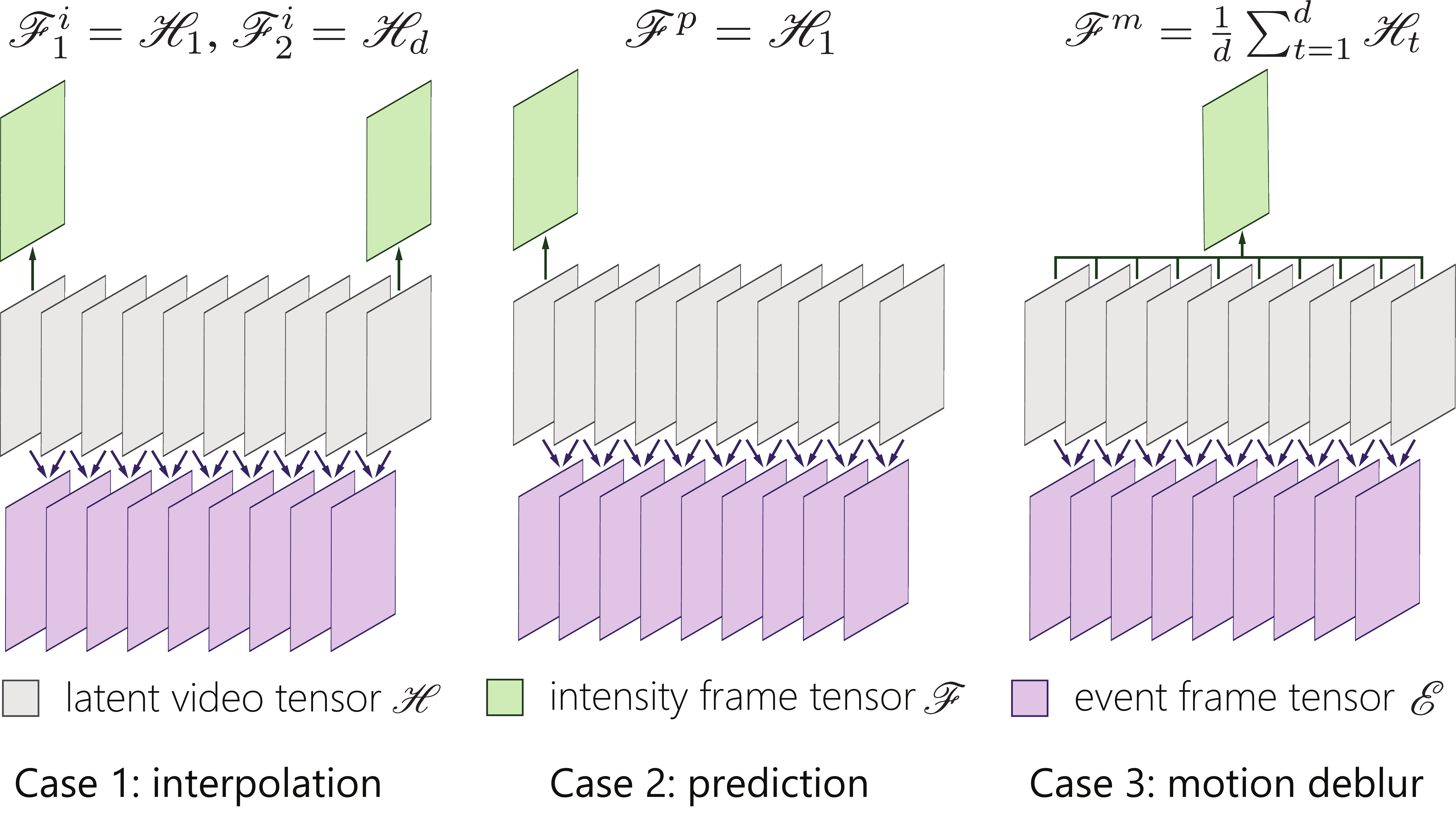}
    \caption{Forward models considered in this paper (See \Sref{ssec:image_formation} for mathematical explanation). Case 1: interpolation from two observed intensity frames and event frames. Case 2: prediction from one observed intensity frame at the beginning and event frames. Case 3: Motion video from a single observed intensity frame and event frames.}
    \label{fig:compare_int}
\end{figure}

\section{Related work}

\customparagraph{Multimodal sensor fusion.} Fusion among different types of sensing modalities for improved quality and functionality is an interesting topic. A related problem to ours is to spatially upsample functional sensors, \eg, depth or hyperspectral sensors, with a high resolution guide image. The fusion problem can be formulated as joint image filtering via bilateral \cite{kopf2007joint}, multi-lateral filters \cite{chan2008noise} or Convolutional Neural Network (CNN) based approach \cite{DJF-ECCV-2016}. For high-speed video sensing, a fusion strategy can be employed between high-speed video cameras (low spatial resolution) and high spatial resolution still cameras (low speed) \cite{bhat2007using, gupta2009enhancing, gupta2010flexible, schubert2008combining, zabrodsky1990attentive}. 

Our paper investigates the temporal upsampling problem. While previous approaches investigate in the framework of compressive sensing \cite{baraniuk2017compressive, he2018computational, iliadis2018deep, llull2013coded, reddy2011p2c2,stankovic2008compressive, wang2017compressive}, we formulate our work as fusing event streams with intensity images to obtain a temporally dense video. Compared to existing literature \cite{Scheerlinck18accv} which integrates event counts per pixel across time, our differentiable model utilizes ``tanh'' functions as event activation units and imposes sparsity constraints on both spatial and temporal domain.

\customparagraph{Event-based image and video reconstruction.}\quad Converting event streams (binary) to multiple-valued intensity frames is a challenging task, yet has been shown beneficial to downstream visual tasks \cite{rebecq2019events}. Existing strategies for image reconstruction include dictionary learning \cite{barua2016direct}, manifold regularization \cite{munda2018real}, optical flow \cite{bardow2016simultaneous}, exponential integration \cite{Pan19cvpr, Scheerlinck18accv}, conditional Generative Adversarial Networks (GAN) \cite{wang2018event} and recurrent neural network \cite{rebecq2019events}. Compared to existing algorithms, our work is the first, to the best of our knowledge, to unify different temporal frame synthesis settings, including interpolation, extrapolation (prediction) and motion deblur (reconstructing a video from a motion-blurred image).

\customparagraph{Non-event-based video frame synthesis.}\quad 1) Interpolation: Early work on video frame interpolation has focused on establishing block-wise \cite{choi2007motion} and/or pixel-wise \cite{krishnamurthy1999frame, luessi2009efficient} correspondences between available frames. Improved performance has been achieved via coarse-to-fine estimation \cite{bergen1992hierarchical}, texture decomposition \cite{wedel2008duality}, and deep neural networks (DNN) \cite{ilg2017flownet}. Recent DNN-based approaches include deep voxel flow \cite{liu2017video}, separable convolution \cite{niklaus2017video}, flow computation and interpolation CNN \cite{jiang2018super}. 2) Prediction: Recent work on future frame prediction has proposed to use adversarial nets \cite{mathieu2015deep}, temporal consistency losses \cite{bhattacharjee2017temporal} and layered cross convolution networks \cite{xue2018visual}. 3) Motion deblur: Recent work on resolving a sharp video/image from blurry image(s) has leveraged adversarial loss \cite{kupyn2018deblurgan}, gated fusion network \cite{zhang2018gated}, ordering-invariant loss \cite{jin2018learning}, \etc.

\section{Approach}
\label{sec:approach}
\subsection{Image formation}
\label{ssec:image_formation}
Assume there exists a high framerate video denoted by tensor $\mathscr{H}\in\mathbb{R}^{h\times w\times d}$, $d > 1$\footnote{$\mathscr{H}$  is indexed on time axis starting from 1. Color channel is omitted here.}. The forward sensing process results in two observational tensors, \ie, the intensity frame tensor $\mathscr{F}$ and event frame tensor $\mathscr{E}$. Our goal is to recover tensor $\mathscr{H}$ based on the observation of intensity and event data.

\customparagraph{Intensity frame tensor.}\quad We consider three sensing cases, \ie 1) interpolation from the first and last frames of $\mathscr{H}$; 2) prediction based on the first frame of $\mathscr{H}$ and 3) motion deblur, in which case the intensity tensor is the summation over time. This can be visualized in \fref{fig:compare_int}. 

\begin{figure}
    \centering
    \begin{subfigure}[b]{0.5\linewidth}
        \includegraphics[width=.95\linewidth]{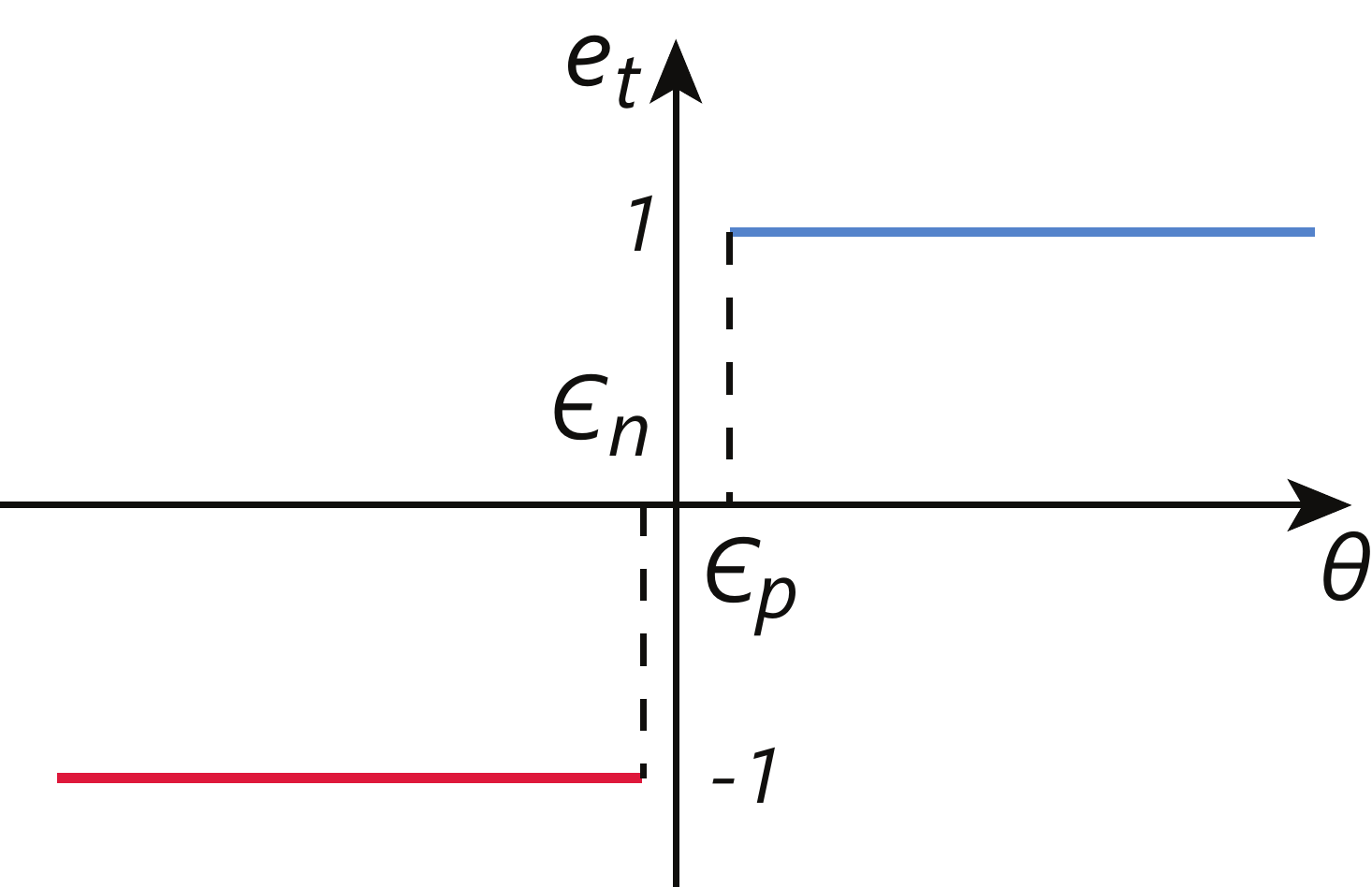}
        \caption{the event firing process}
        \label{fig:ev_thres}
    \end{subfigure}%
    \begin{subfigure}[b]{0.5\linewidth}
        \includegraphics[width=.95\linewidth]{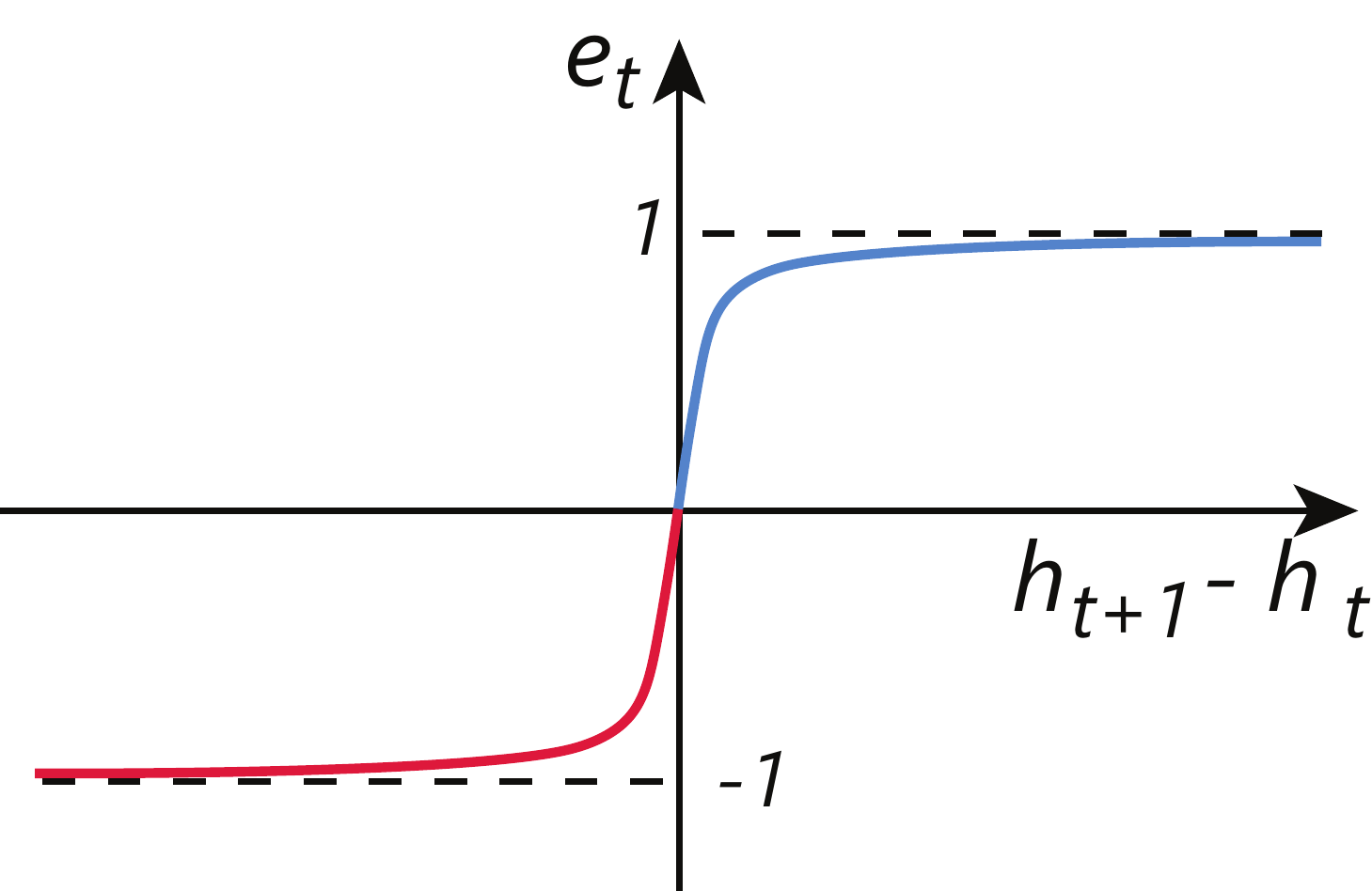}
        \caption{our proposed activation unit}
        \label{fig:ev_tanh}
    \end{subfigure}%
    \caption{Comparison of the event firing process and our proposed differentiable model. $h_t$ denotes a pixel of $\mathscr{H}_t$.}
    \label{fig:activation}
\end{figure}

\customparagraph{Event frame tensor.}\quad As previously introduced, a pixel fires a binary output/event if the log-intensity changes beyond a threshold (positive or negative). This thresholding model can be viewed in \fref{fig:ev_thres}. Mathematically, the event firing process can be expressed as,

\begin{equation}
    e_t = \left\{\begin{array}{cc}
        1 & \theta > \epsilon_p \\
        -1 & \theta < -\epsilon_n\\
        0 & otherwise
    \end{array}\right.,
    \label{eq:event}
\end{equation}
where $\theta = \log(I_t + b) - \log(I_0 + b)$. If $e_t=0$, no events are generated. In order to approximate this event firing process, we model each event frame as a function of the adjacent frames from the high framerate tensor $\mathscr{H}$, \ie,

\begin{equation}
    \mathscr{E}_{t} = \tanh\Big\{\alpha \big[\mathscr{H}_{t+1} - \mathscr{H}_{t}\big]\Big\},
    \label{eq:tanh}
\end{equation}
where $\alpha$ is a tuning parameter to adjust the slope of the activation curve. This function can be viewed in \fref{fig:ev_tanh}. Based on this formulation, a video tensor with $d$ temporal frames correspond to $d-1$ event frames.

\begin{figure*}
    \centering
    \begin{subfigure}[t!]{0.2\linewidth}
        \includegraphics[width=.97\linewidth]{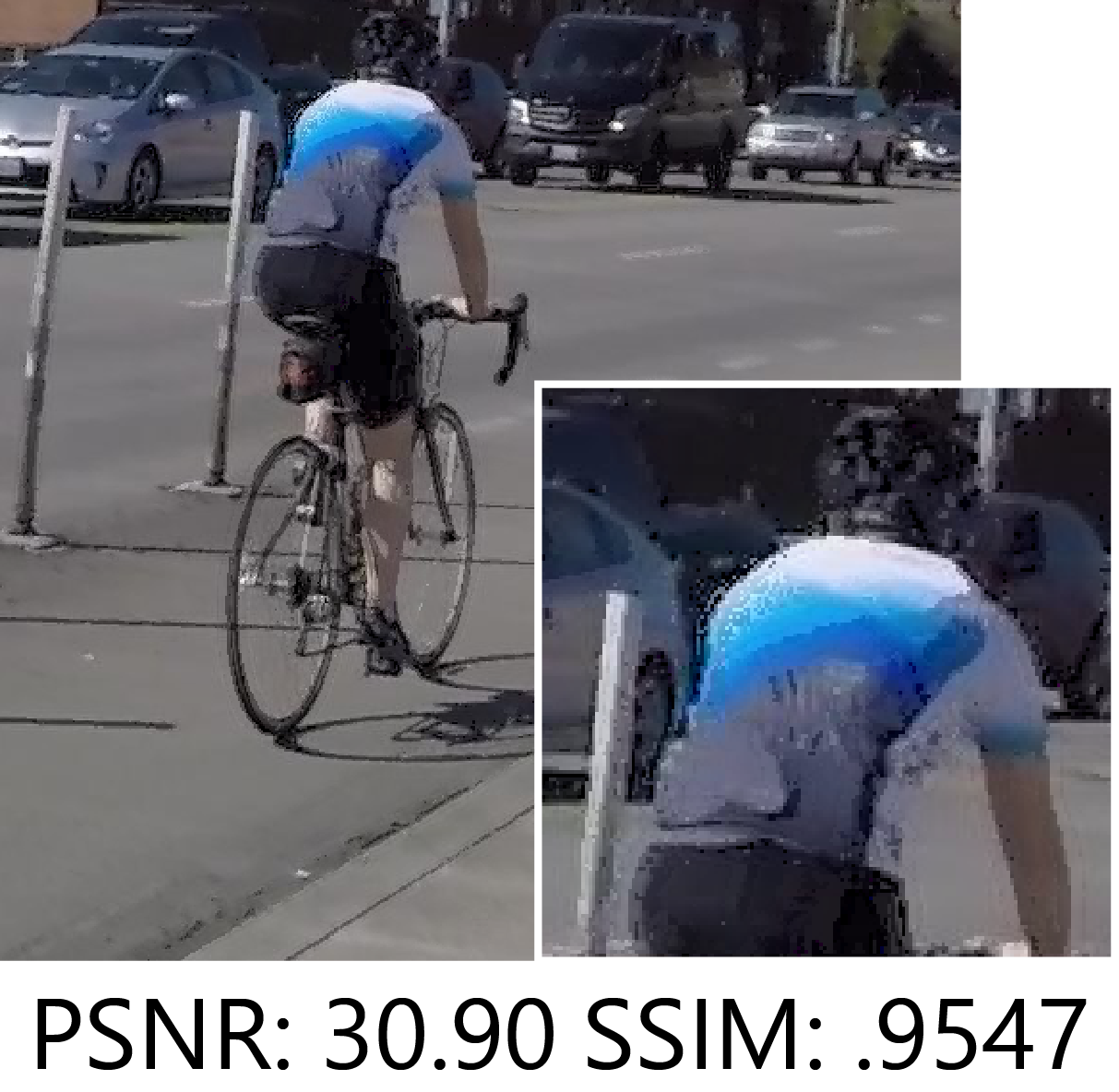}
        \caption{$\mathcal{L}_{pix}$}
        \label{fig:loss_pix}
    \end{subfigure}%
    \begin{subfigure}[t!]{0.2\linewidth}
        \includegraphics[width=.97\linewidth]{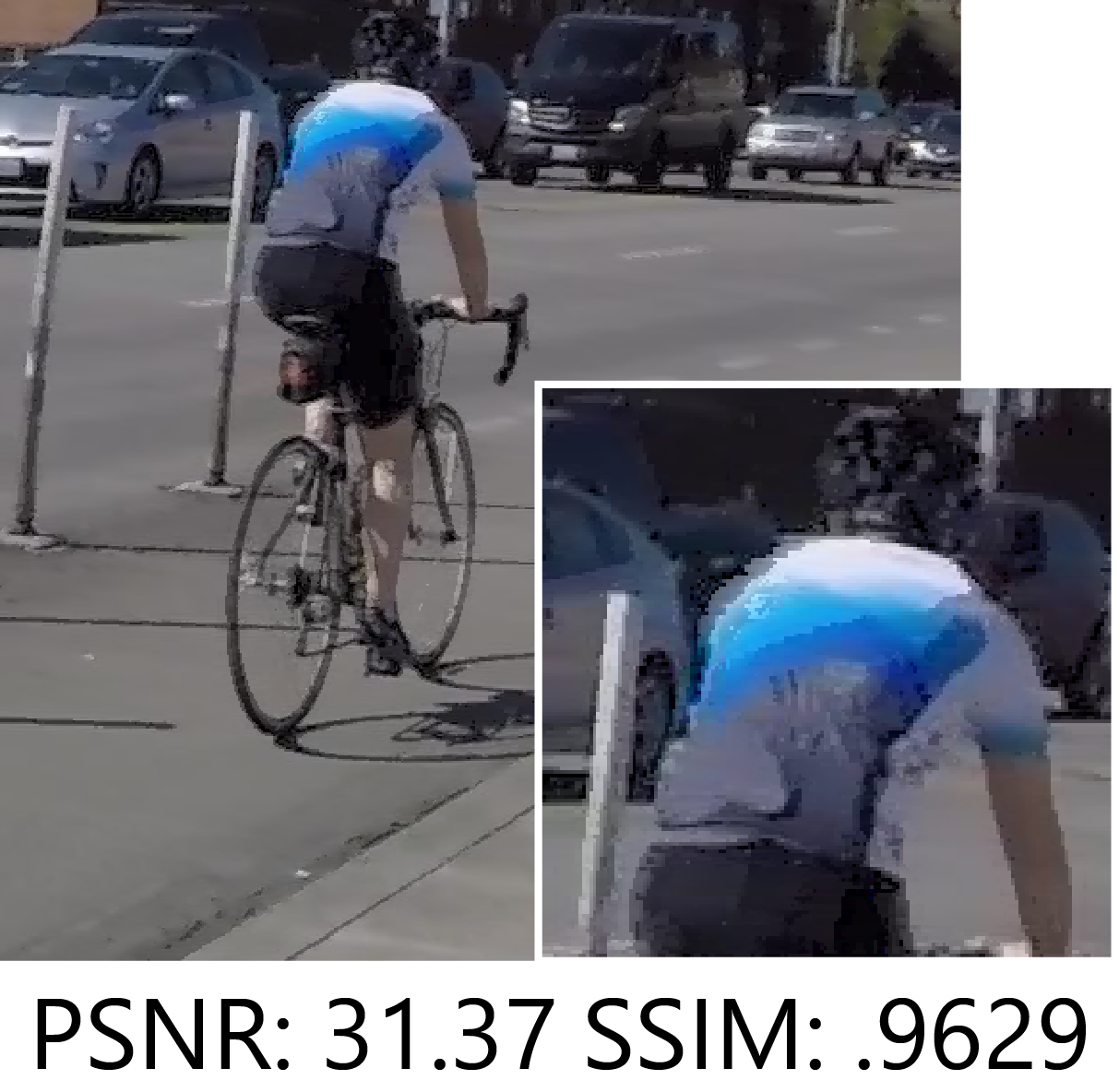}
        \caption{$\mathcal{L}_{pix}+\lambda_t\mathcal{L}_{TV_t}$}
        \label{fig:loss_tvt}
    \end{subfigure}%
    \begin{subfigure}[t!]{0.2\linewidth}
        \includegraphics[width=.97\linewidth]{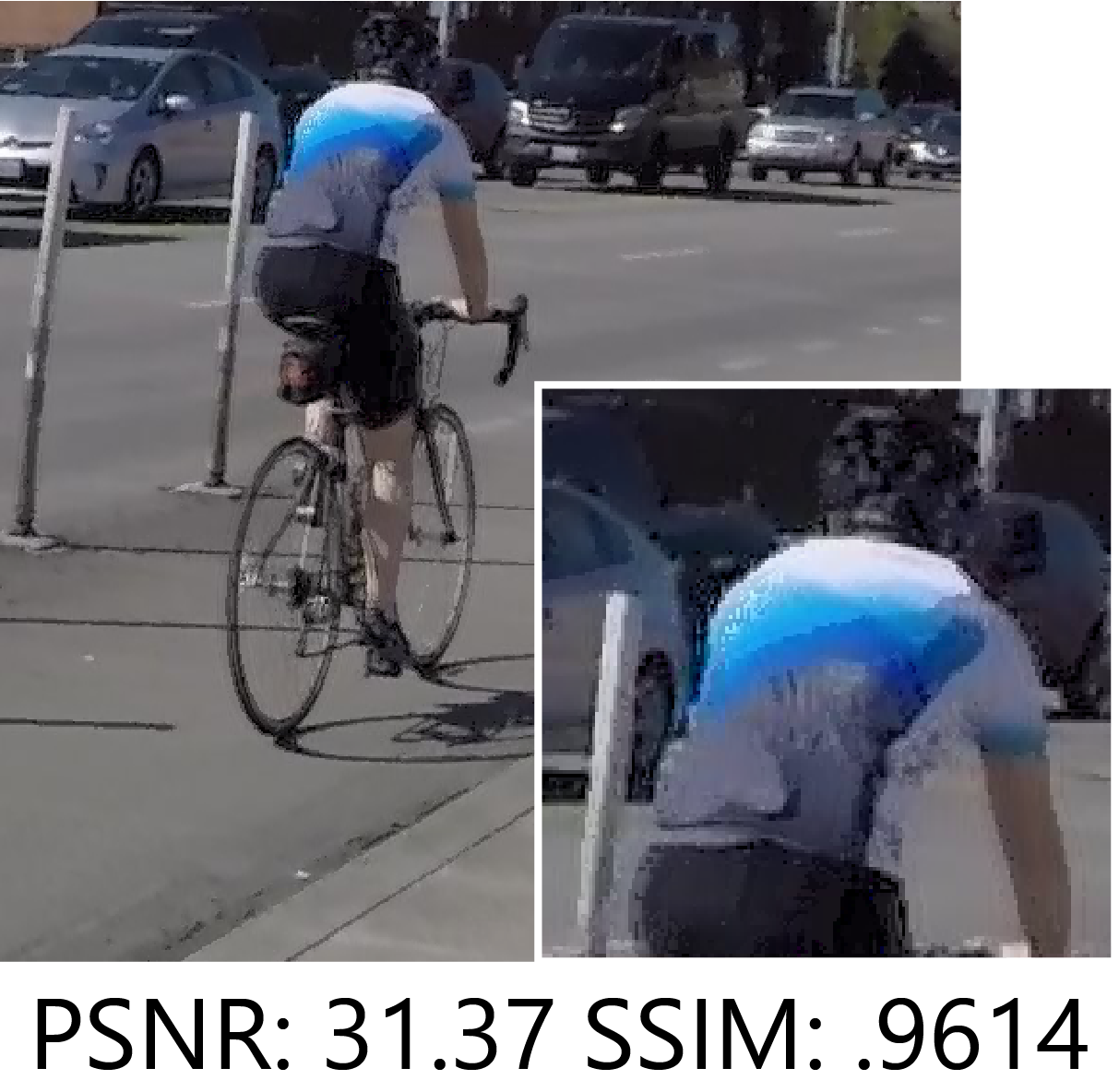}
        \caption{$\mathcal{L}_{pix}+\lambda_{xy}\mathcal{L}_{TV_{xy}}$}
        \label{fig:loss_tvxy}
    \end{subfigure}%
    \begin{subfigure}[t!]{0.2\linewidth}
        \includegraphics[width=.97\linewidth]{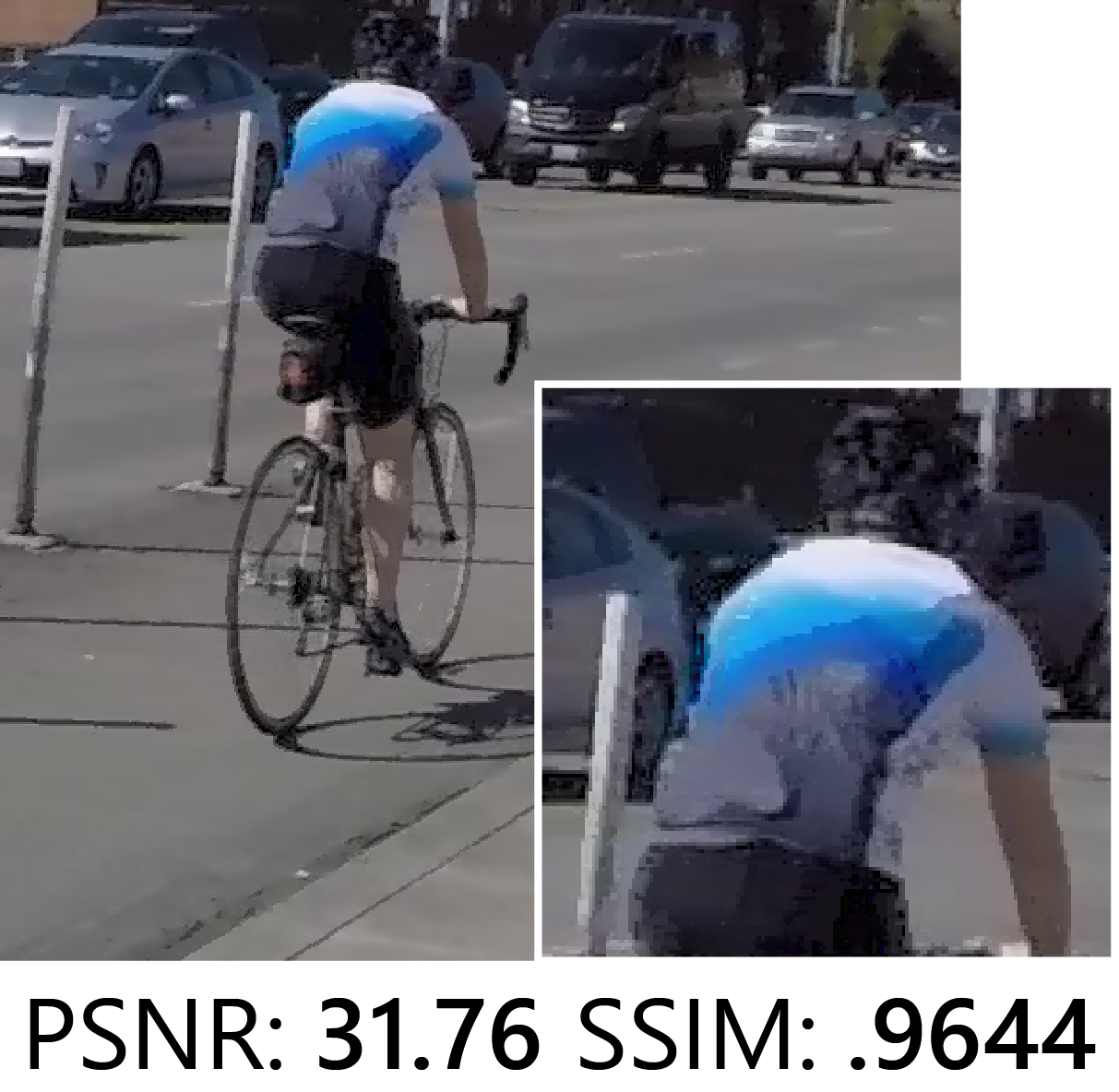}
        \caption{$\mathcal{L}_{pix}+\mathcal{L}_{TV}$}
        \label{fig:loss_tvxyt}
    \end{subfigure}%
    \begin{subfigure}[t!]{0.2\linewidth}
        \includegraphics[width=.97\linewidth]{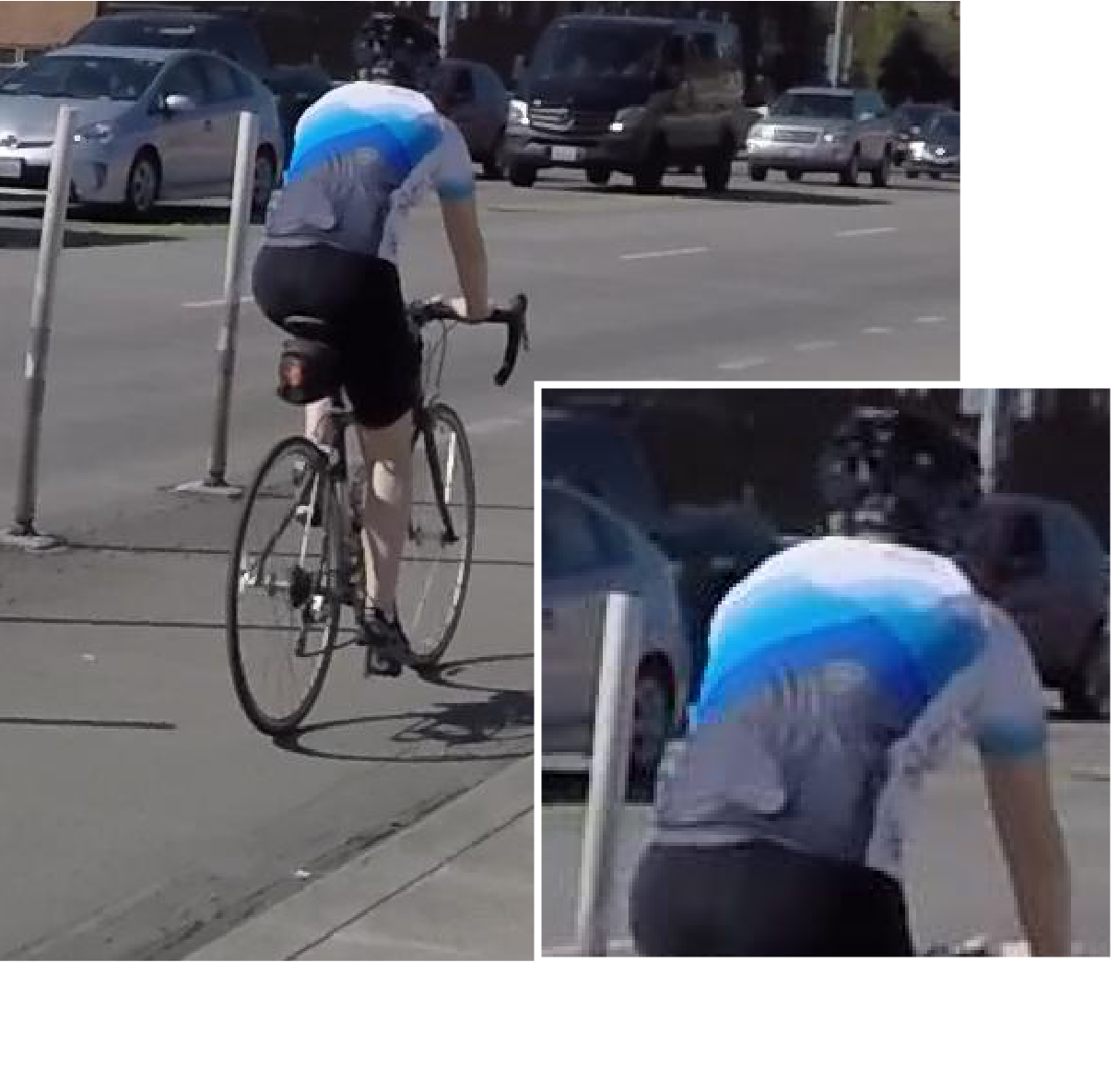}
        \caption{ground truth}
        \label{fig:loss_gt}
    \end{subfigure}%
    \caption{Comparison of different loss functions (simulated single-frame interpolation). $\mathcal{L}_{TV}=\lambda_t\mathcal{L}_{TV_t}+\lambda_{xy}\mathcal{L}_{TV_{xy}}$. }
    \label{fig:losses}
\end{figure*}

\subsection{Differentiable model-based reconstruction}
The DMR is performed by minimizing a weighted combination of several loss functions. The objective function is formed as,

\begin{dmath}
    \label{eq:objective}
    \hat{\mathscr{H}} = \argmin_{\mathscr{H}}\mathcal{L}_{pix}(\mathscr{H}, \mathcal{F}, \mathcal{E}) + \mathcal{L}_{TV}(\mathscr{H}) 
\end{dmath}

\customparagraph{Pixel loss.}\quad The pixel loss includes per-pixel difference loss against intensity and event pixels in $\ell_1$ norm, \ie,

\begin{equation}
\begin{split}
    \mathcal{L}_{pix}(\mathscr{H},\mathcal{F},\mathcal{E}) & = \mathbb{E}_{fpix}\big[\norm{\mathcal{F} - \mathcal{A}(\mathscr{H})}_1\big] \\
    & + \lambda_{e}\mathbb{E}_{epix}\big[\norm{\mathcal{E} - \mathcal{B}(\mathscr{H})}_1\big],
    \label{eq:pix_loss}
\end{split}
\end{equation}
over the entire available data range. $\mathcal{F}$ and $\mathcal{E}$ denote the captured intensity and event data, respectively. $\mathcal{A}$ and $\mathcal{B}$ denote the forward sensing models described in \fref{fig:compare_int} and \Eref{eq:tanh}. $\mathbb{E}_x$ represents expectation with respect to the observed pixels/events.

\customparagraph{Sparsity loss.}\quad We employ total variation (TV) sparsity in the spatial and temporal dimensions of the high-res tensor $\mathscr{H}$. The TV sparsity loss is defined as: 
\begin{dmath}
    \mathcal{L}_{TV}(\mathscr{H}) = \lambda_{xy}\mathbb{E}_{hpix}\Big[\norm{\dot{\mathscr{H}}_{xy}}_1\Big] + \lambda_{t}\mathbb{E}_{hpix}\Big[\norm{\dot{\mathscr{H}}_t}_1\Big],
    \label{eq:tv_loss}
\end{dmath}
where $\dot{\mathscr{H}}_{xy}=\pdv{\mathscr{H}}{x}+\pdv{\mathscr{H}}{y}$ and $\dot{\mathscr{H}}_t = \pdv{\mathscr{H}}{t}$. We later denote $\mathcal{L}_{TV_{xy}} = \mathbb{E}_{hpix}\big[\norm{\dot{\mathscr{H}}_{xy}}_1\big]$ and $\mathcal{L}_{TV_{t}} = \mathbb{E}_{hpix}\big[\norm{\dot{\mathscr{H}}_{t}}_1\big]$. $\mathcal{L}_{TV_{xy}}$ can be viewed as a denoising term for intensity tensor, and $\mathcal{L}_{TV_{t}}$ can be viewed as an event denoising term. A comparison of the performance for each loss function is shown in \fref{fig:losses}. The figure shows a synthetic case for single-frame interpolation. We use three frames, resulting in two event frames (\Eref{eq:event}). Combining the spatial and temporal TV losses resullts in better performance.

\begin{figure}
    \centering
        \includegraphics[width=\linewidth]{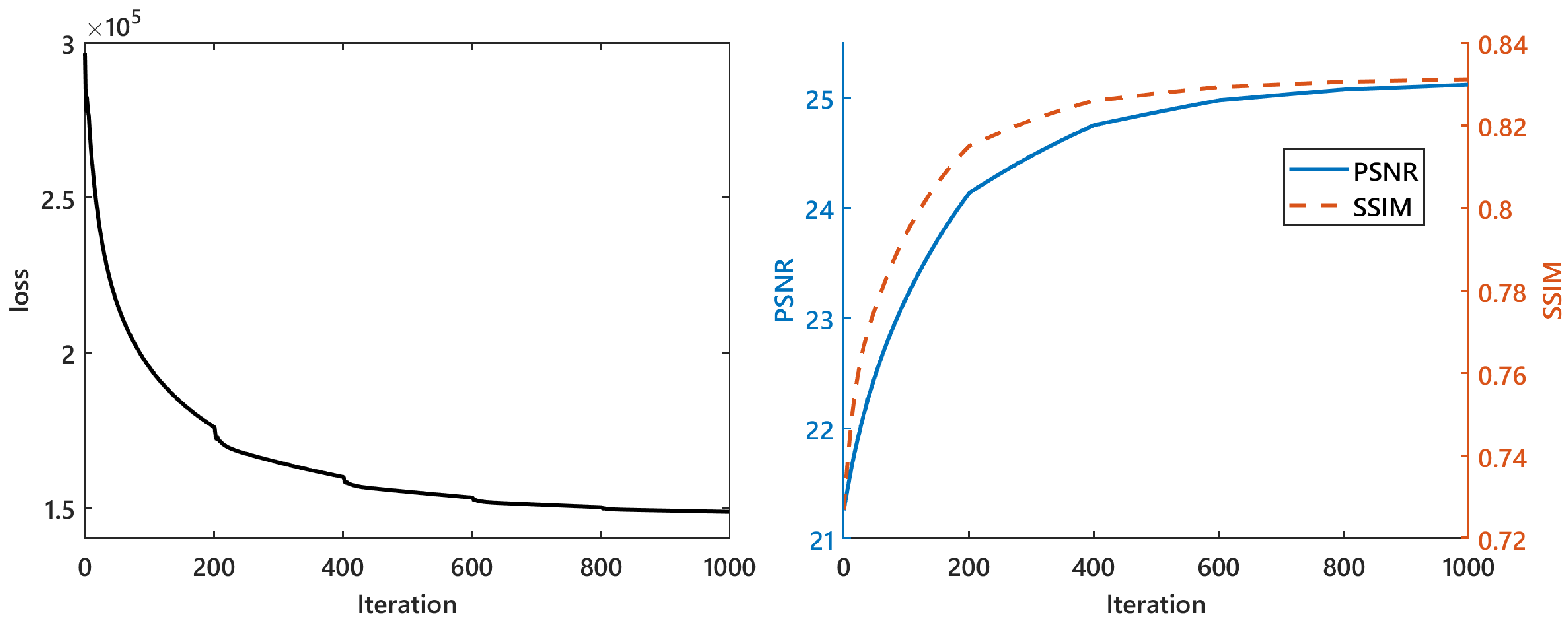}
        \label{fig:loss_progress}
    \caption{Loss values and accuracy (PSNR and SSIM) during DMR optimization.} 
    \label{fig:opt}
\end{figure}

\customparagraph{Implementation.}\quad We use stochastic gradient descent to optimize \Eref{eq:objective} so as to reconstruct the latent high-res tensor. Our algorithm is implemented in TensorFlow. We use Adam optimizer. The learning rate varies depending on the tensor size as well as related parameters. Empirically, we recommend 0.002 as initial value. We recommend to schedule the learning rate to decrease $5\times$ every 200 epochs. The momenta $\beta_1=0.9, \beta_2=0.99$. For the case of interpolation, we initialize the high-res tensor $\mathscr{H}$ by linearly blending the two available low-res frames. For prediction and motion deblur, we initialize the high-res tensor using the available single low-res frame. An example of the optimization progress can be viewed in \fref{fig:opt}. As the loss decreases, both PSNR and SSIM increase and gradually converge.

\subsection{Binning events into event frames}
\label{ssec:binning}
Our event sensing model requires binning events into frames. The ideal binning strategy would be ``one frame per event''. However, this binning strategy is unnecessarily expensive. For example, the events between two consecutive frames (22 FPS in \cite{mueggler2017event}) may vary from thousands to tens of thousands, resulting in computational challenges and redundancy. However, events happening at different locations but at very close timestamps can be processed in the same event frame. Therefore, we design and compare two binning strategies:

\customparagraph{Binning 1 (proposed):}\quad For an incoming event, if its spatial location already has an event in the current event frame, then cast it into a new event frame; otherwise, this incoming event will stay in the current event frame. In this case, each event frame should only have three values, \ie, \{-1, 0, 1\}.

\customparagraph{Binning 2:}\quad Similar to several previous work \cite{bardow2016simultaneous, barua2016direct, rebecq2019events, wang2018event}, where events are stacked/integrated over a time window, we allow each event frame to have more than three values. However, since the ``tanh'' function in \Eref{eq:tanh} only outputs values between -1 and 1, we modify our event sensing model to have a summation operation over several sub-event frames. Mathematically, $\mathscr{E}_{b2} = \sum_t{\mathscr{E}_t}$. 

\begin{figure}
    \centering
    \begin{subfigure}[b]{0.5\linewidth}
        \includegraphics[width=.95\linewidth]{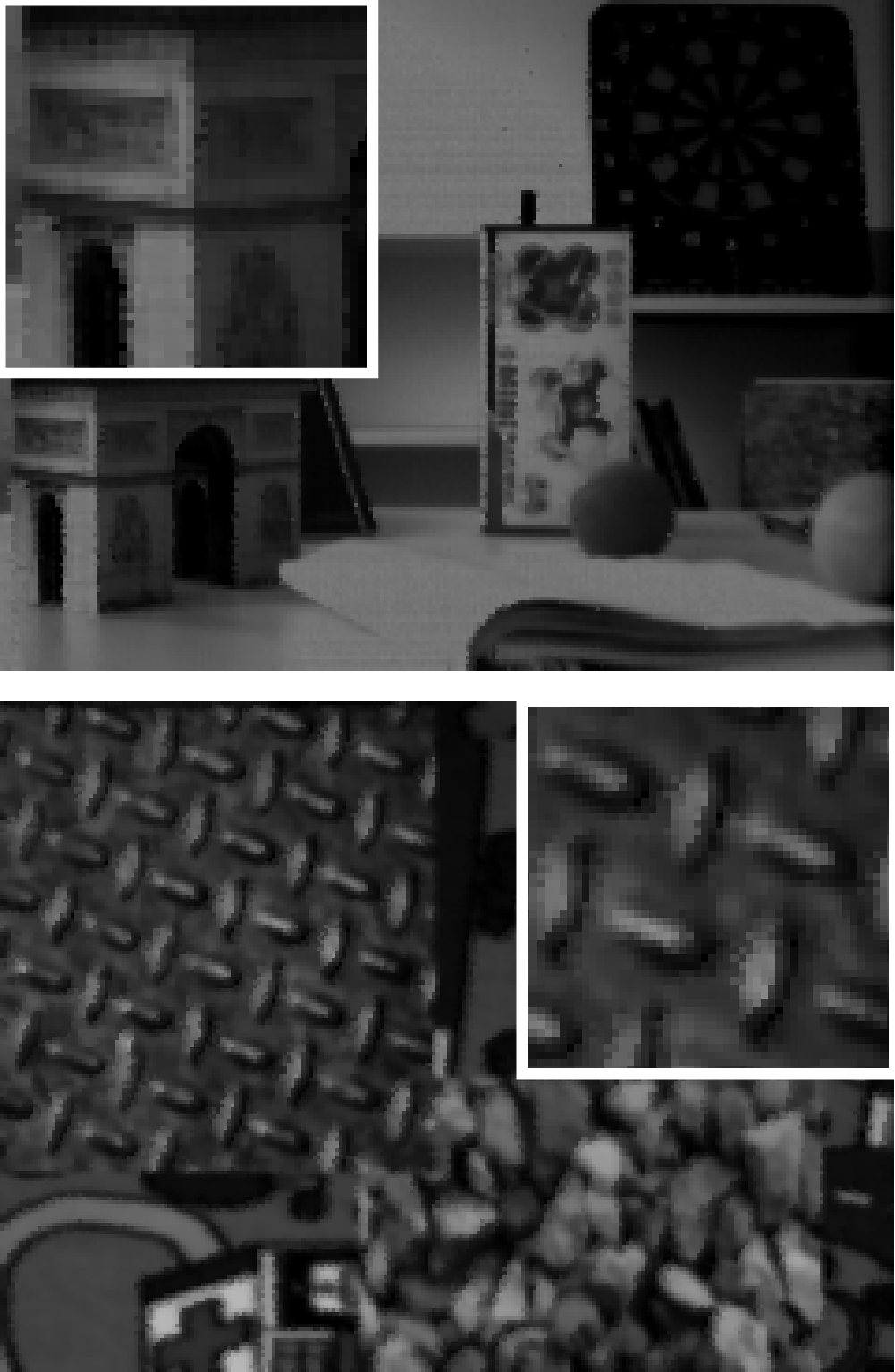}
        \caption{Binning 1}
        \label{fig:binning_1}
    \end{subfigure}%
    \begin{subfigure}[b]{0.5\linewidth}
        \includegraphics[width=.95\linewidth]{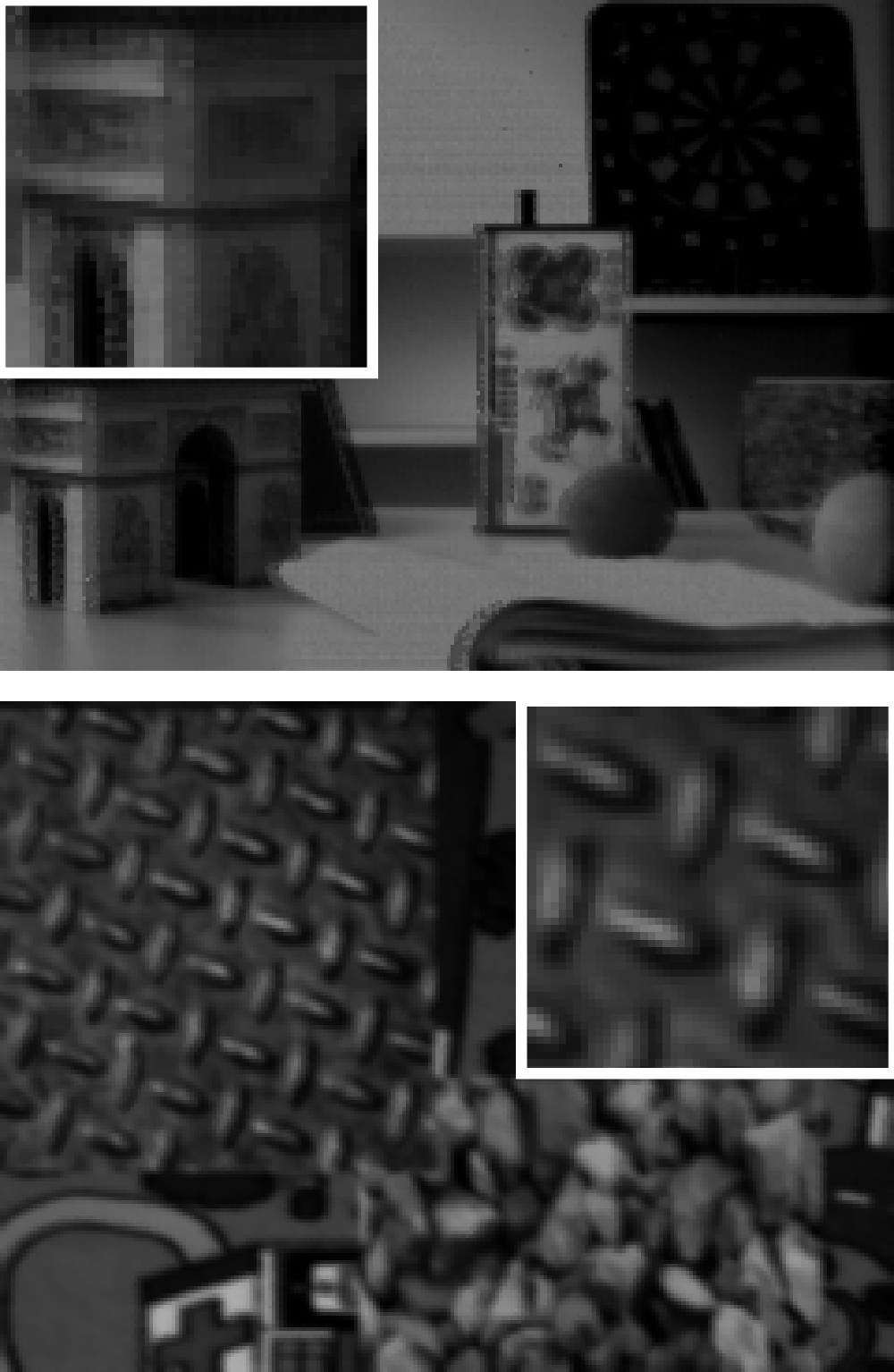}
        \caption{Binning 2}
        \label{fig:binning_2}
    \end{subfigure}%
    \caption{Comparison of two binning strategies applied to frame interpolation using the DAVIS dataset.}
    \label{fig:binning}
\end{figure}

We show DMR results for a frame interpolation case using DAVIS dataset \cite{mueggler2017event} in  \fref{fig:binning}. We use two consecutive intensity frames and the events in-between. In Row 1 (``slider_depth''), 9 event frames are binned from over 7, 700 events using Binning 1. Row 2 (``simulation_3_planes'') has 19 event frames from over 40, 000 events. For Binning 2, we match the sub-event frame number with Binning 1 so as to compare the performance. Frame \#2 is shown. Our results show that Binning 1 preserves sharp spatial structures\footnote{A more detailed analysis and complete slow motion videos can be found in the supplementary material.}. For subsequent experiments, we use Binning 1.

\subsection{Learning a residual denoiser}\label{subsec:rd}
Although our proposed DMR can handle a variety of fusion settings, we observe that the DMR results may have visual artifacts. This is due to the ill-posedness of the fusion problem and different noise levels between the two sensing modalities. In order to address these issues, we model the artifacts outcome of DMR as additive ``noise'' and propose a ``denoising'' process to remove the artifacts. Inspired by ResNet \cite{he2016deep} and DnCNN \cite{zhang2017beyond}, we employ the residual learning scheme and train a residual denoiser (RD).  Rather than training the denoiser from various levels of artificial noise, we design to train the network from the outcome of DMR. Mathematically, the residual $\mathscr{R}$ is expressed as,
\begin{equation}
    \mathscr{R} = \hat{\mathscr{H}} - \mathscr{H}_g,
\end{equation}
where $\hat{\mathscr{H}}$ represents the reconstructed frame from DMR, and $\mathscr{H}_g$ represents the ground truth frame. We use a residual block similar to \cite{zhang2018ffdnet}, which has a \{conv + ReLU\} and a \{conv\} layer at the beginning and end, with 17 intermediate layers of \{conv + BN + ReLU\}. The kernel size is $3\times3$ with stride of 1. The loss function for our denoiser is the mean squared error of $\hat{\mathscr{H}}$ and $\mathscr{R}$. During training, we augment data by randomizing the configuration parameters (including the running epochs) in DMR, summarized in \Tref{tab:aug}. The goal of this augmentation is 1) to prevent overfitting; 2) to enforce learning of our DMR process; 3) to alleviate effects due to non-optimal parameter tunning. Our denoiser is single-frame, as we seek to enhance each DMR output frame iteratively \emph{without} comprimising the variety of DMR fusion settings.

\begin{table}
    \caption{Augmentation recipe}\label{tab:aug}
    \centering
    \begin{tabular}{| c | c | c |}
    \hline
    source & notation & value range \\
    \hline\hline
    Eq. \eqref{eq:event} & $\epsilon_p$, $\epsilon_n$  & (0, 0.05) \\
    Eq. \eqref{eq:tanh} & $\alpha$ & (8, 20) \\
    Eq. \eqref{eq:pix_loss} & $\lambda_e$ & (0.1, 0.5) \\
    Eq. \eqref{eq:tv_loss} & $\lambda_{xy}$ & (0.3, 0.8) \\
    Eq. \eqref{eq:tv_loss} & $\lambda_t$ & (0.2, 0.6) \\
    \hline\hline
    ~ & event percentage & (0\%, 20\%) \\
    ~ & PBR learning rate & (0.001, 0.009) \\
    ~ & PBR epoch(s) & (1, 350) \\
    \hline
    \end{tabular}
    
\end{table}

\section{Experiment results}\label{sec:exp}
We design several experiments to show the effectiveness of our algorithm. For DMR, we evaluate the three cases described in \fref{fig:compare_int} on the DAVIS dataset \cite{mueggler2017event}, and compare against state-of-the-art event-based algorithms, \ie, Complementary Filter \cite{Scheerlinck18accv} and Event-based Double Integral \cite{Pan19cvpr}. For RD, we evaluate the effectiveness of our learning strategy by comparing with Gaussian denoisers, \eg, DnCNN \cite{zhang2017beyond} and FFDNet \cite{zhang2018ffdnet}. We finally compare our results with a non-event-based frame interpolation algorithm, SepConv \cite{niklaus2017video}.

\subsection{Results for DMR}\label{ssec:dmr_results}
\begin{figure*}
    \centering
    \begin{subfigure}[b]{0.16\linewidth}
        \includegraphics[width=.96\linewidth]{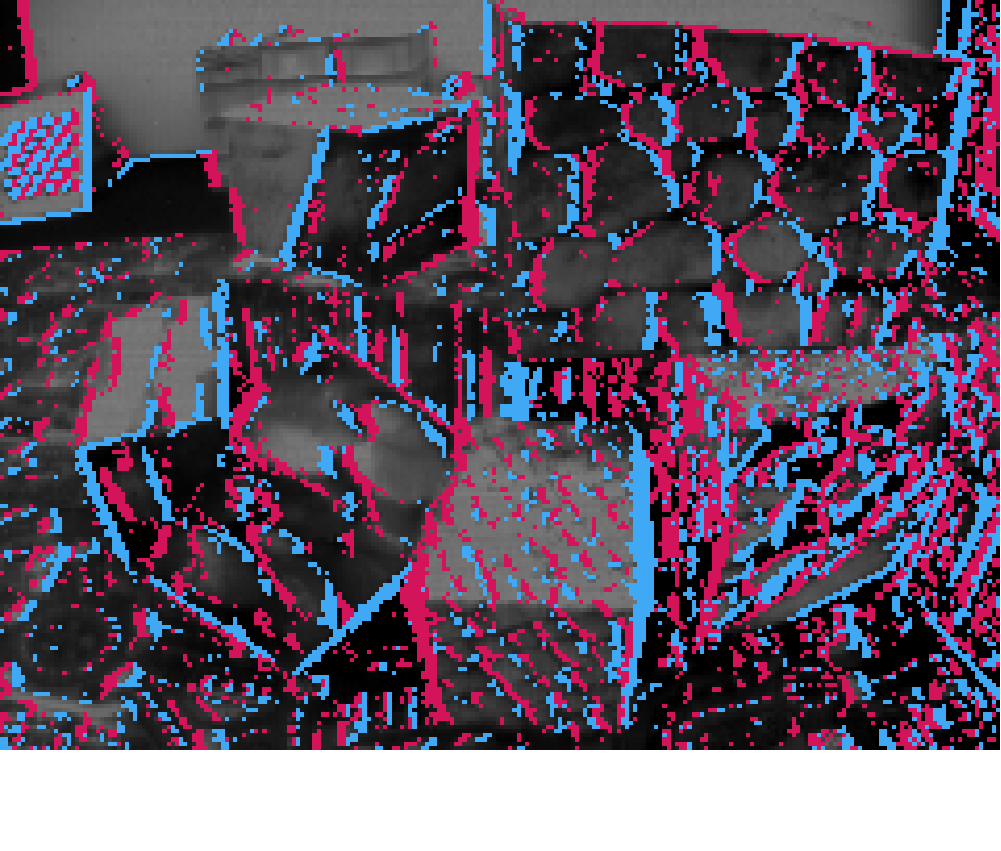}
        \caption{Frame \#1 + events}
        \label{fig:interp_f1_ev}
    \end{subfigure}%
    \begin{subfigure}[b]{0.16\linewidth}
        \includegraphics[width=.96\linewidth]{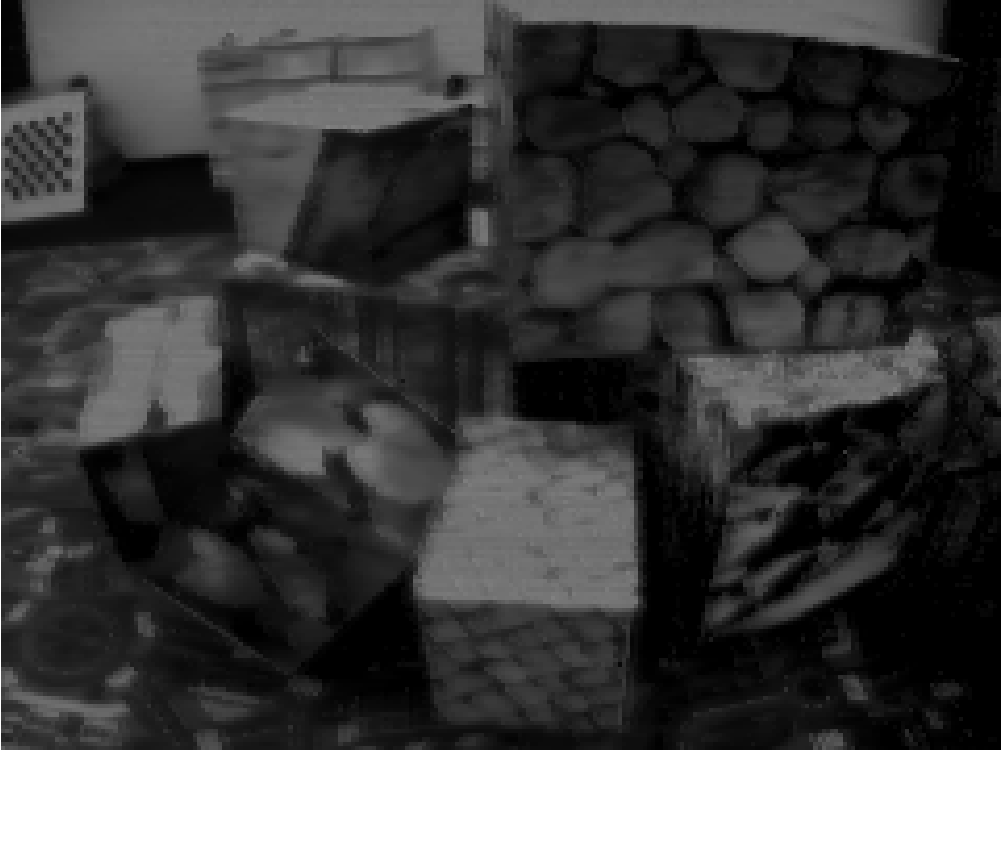}
        \caption{Frame \#5}
        \label{fig:interp_f5}
    \end{subfigure}%
    \begin{subfigure}[b]{0.16\linewidth}
        \includegraphics[width=.96\linewidth]{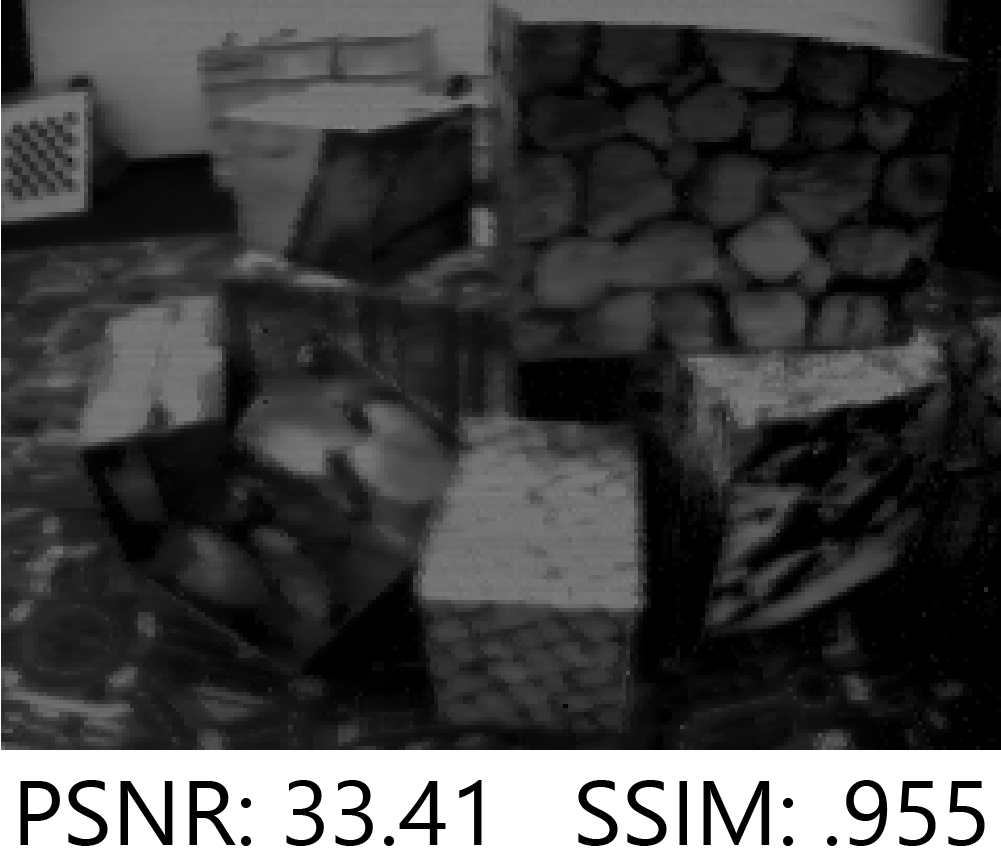}
        \caption{Frame \#10}
        \label{fig:interp_f10}
    \end{subfigure}%
    \begin{subfigure}[b]{0.16\linewidth}
        \includegraphics[width=.96\linewidth]{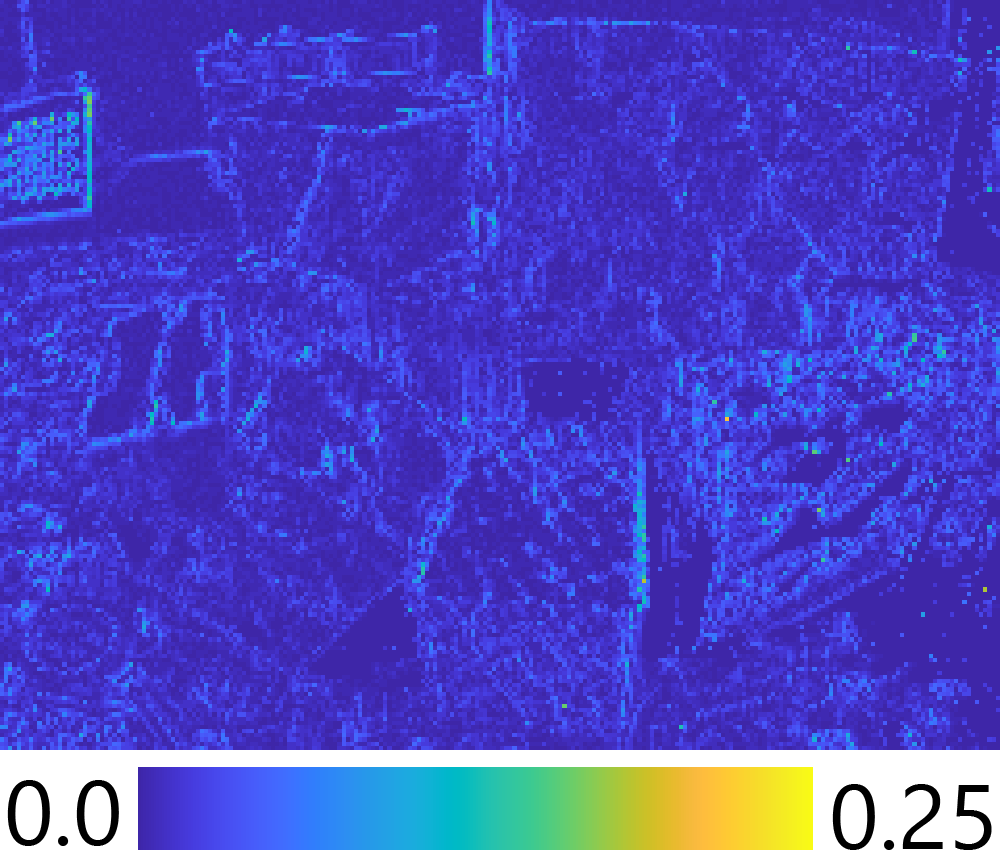}
        \caption{Error map \#10}
        \label{fig:interp_f10}
    \end{subfigure}%
    \begin{subfigure}[b]{0.16\linewidth}
        \includegraphics[width=.96\linewidth]{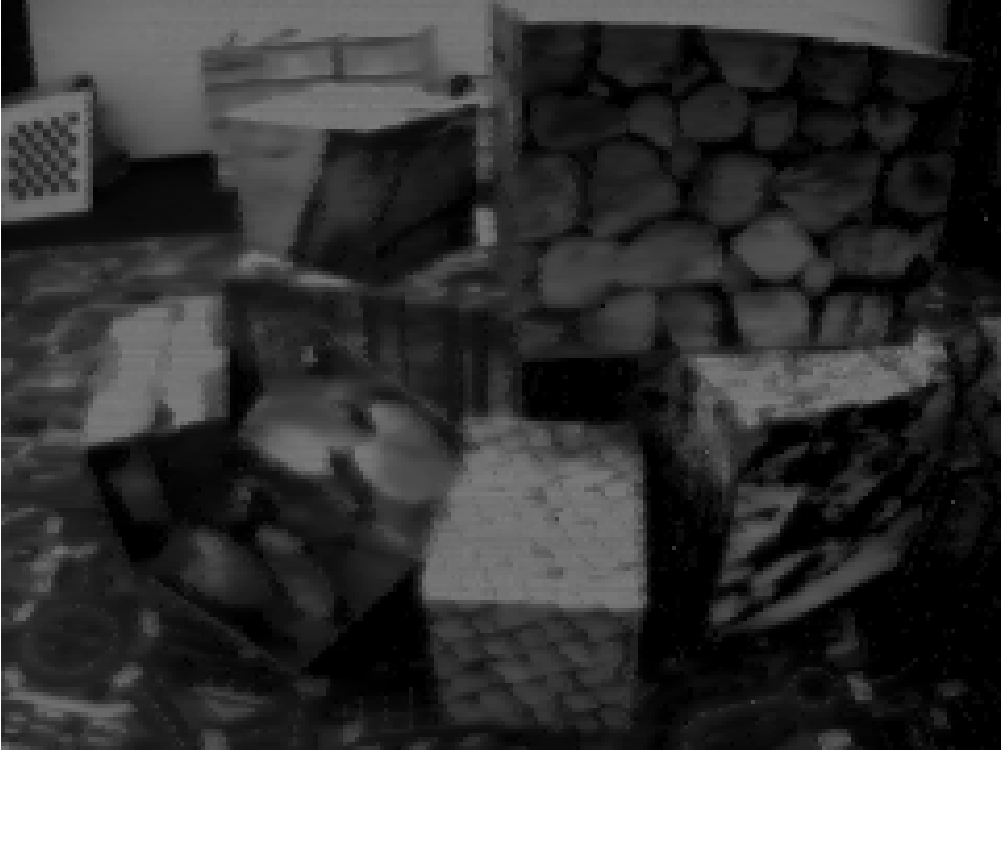}
        \caption{Frame \#16}
        \label{fig:interp_f16}
    \end{subfigure}%
    \begin{subfigure}[b]{0.16\linewidth}
        \includegraphics[width=.96\linewidth]{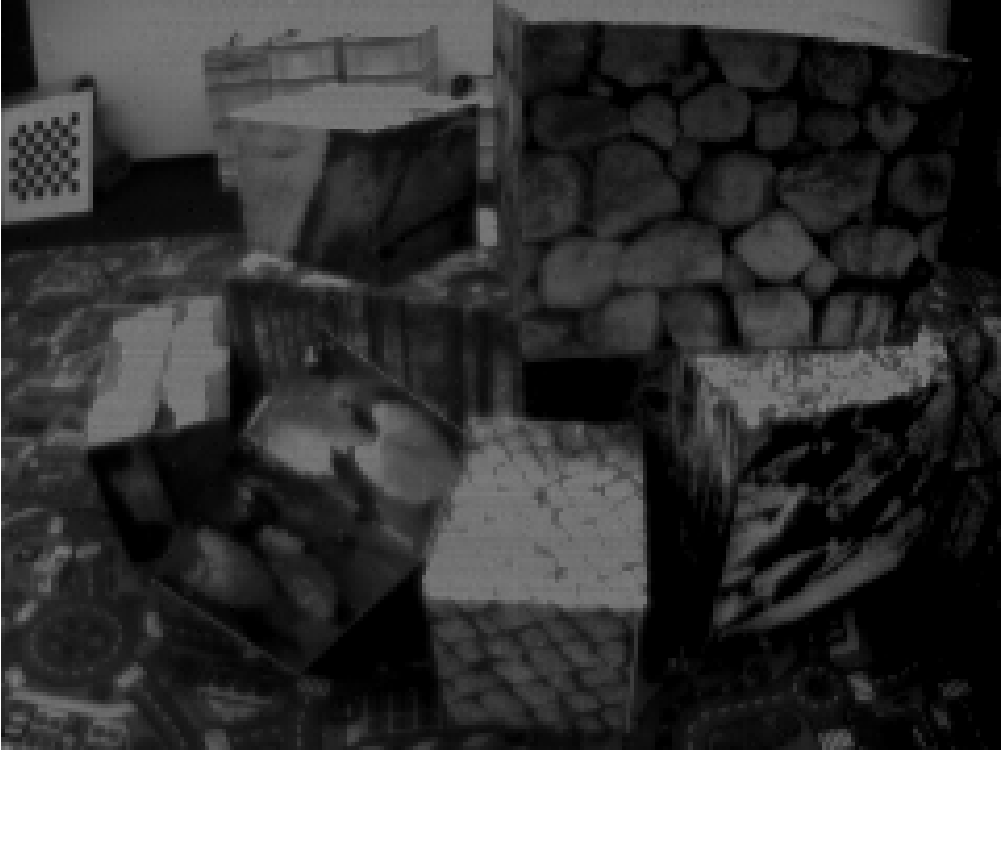}
        \caption{Frame \#21}
        \label{fig:interp_f3}
    \end{subfigure}%
    \caption{Frame interpolation. The start and end frames, as well as in-between events, are used as input. Frame \#10 is compared against the ground truth middle frame.}
    \label{fig:interp_davis}
\end{figure*}

\customparagraph{Interpolation.}\quad We first show interpolation results in \fref{fig:interp_davis}. We use three consecutive frames from \cite{mueggler2017event}, withholding the middle frame. The intermediate events bin into 20 event frames. The ground truth middle frame is the closest to Frame \#10.

\begin{figure*}
    \centering
    \begin{subfigure}[b]{0.16\linewidth}
        \includegraphics[width=.96\linewidth]{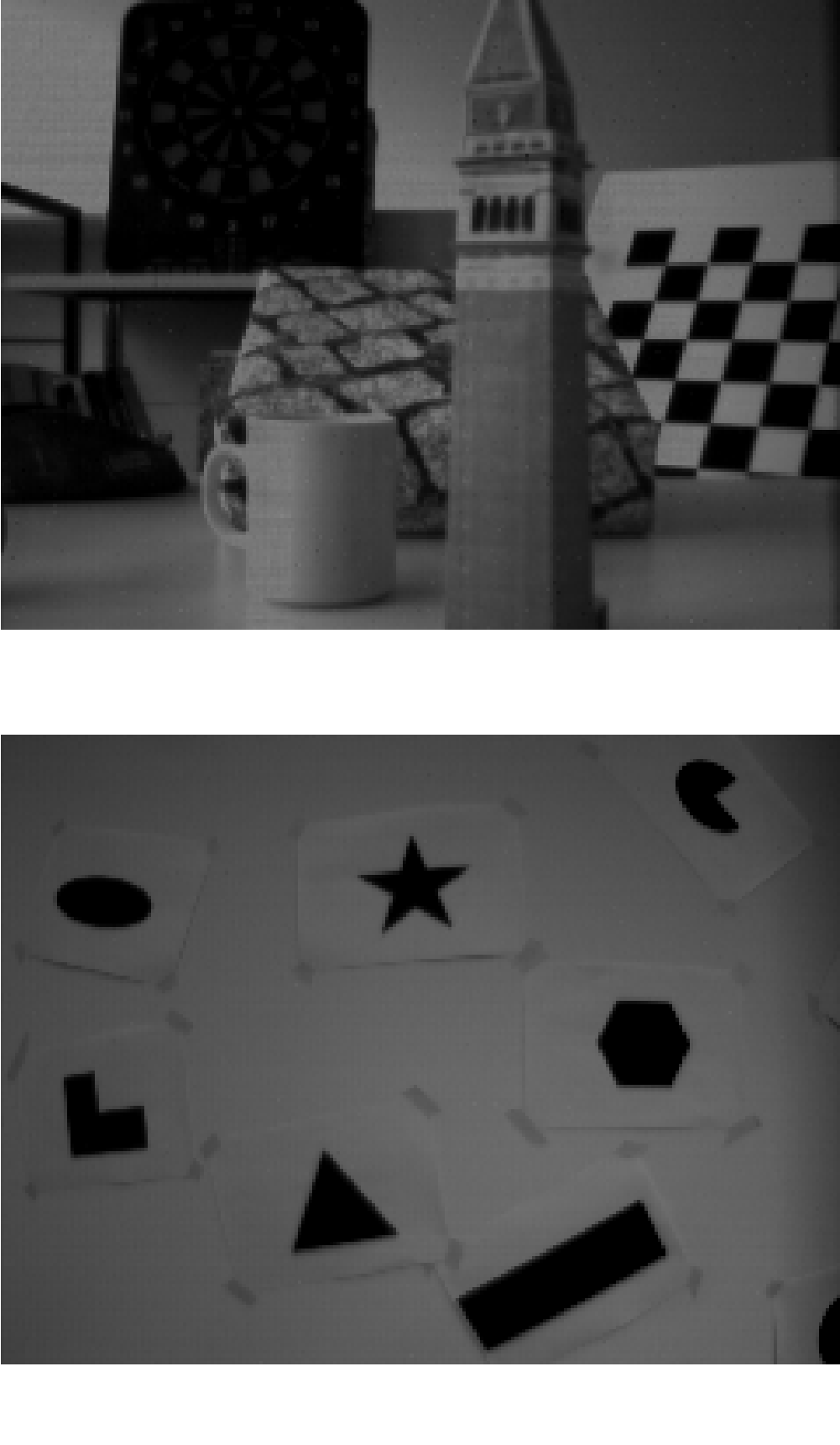}
        \caption{Start frame}
        \label{fig:pred_f1}
    \end{subfigure}%
    \begin{subfigure}[b]{0.16\linewidth}
        \includegraphics[width=.96\linewidth]{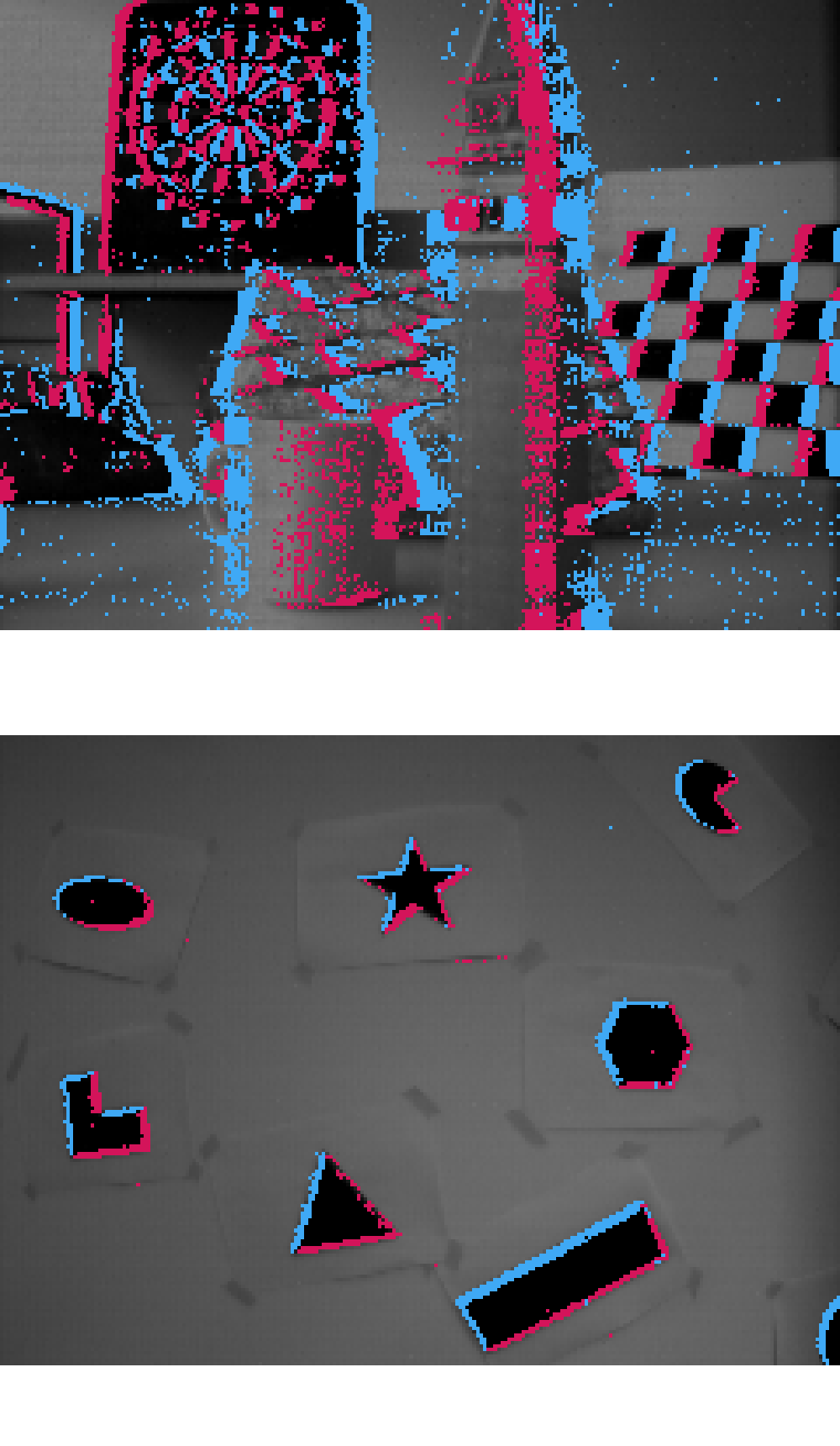}
        \caption{w/ future events}
        \label{fig:pred_f1_ev}
    \end{subfigure}%
    \begin{subfigure}[b]{0.16\linewidth}
        \includegraphics[width=.96\linewidth]{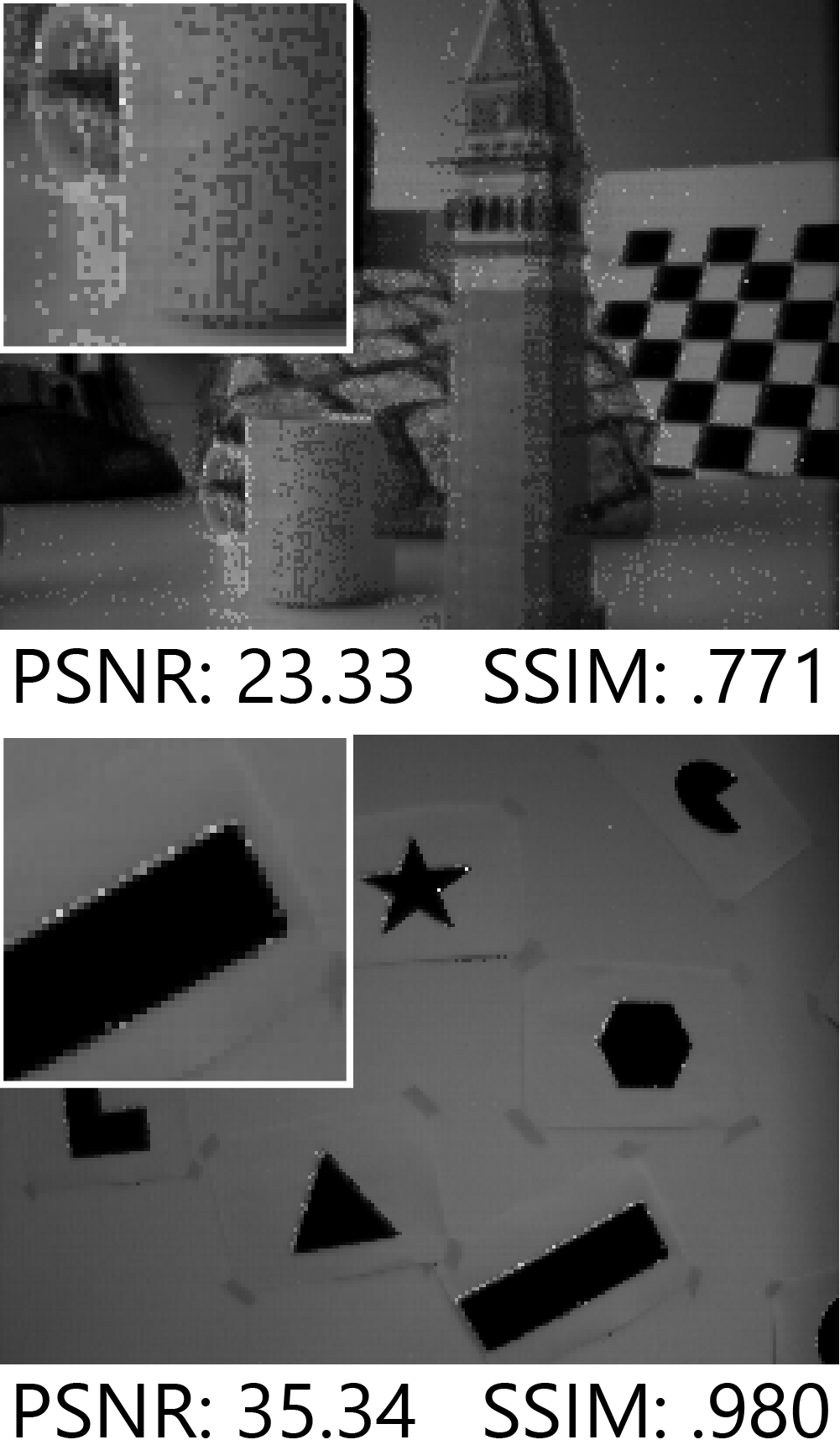}
        \caption{CF \cite{Scheerlinck18accv}}
        \label{fig:pred_cf_rec}
    \end{subfigure}%
    \begin{subfigure}[b]{0.16\linewidth}
        \includegraphics[width=.96\linewidth]{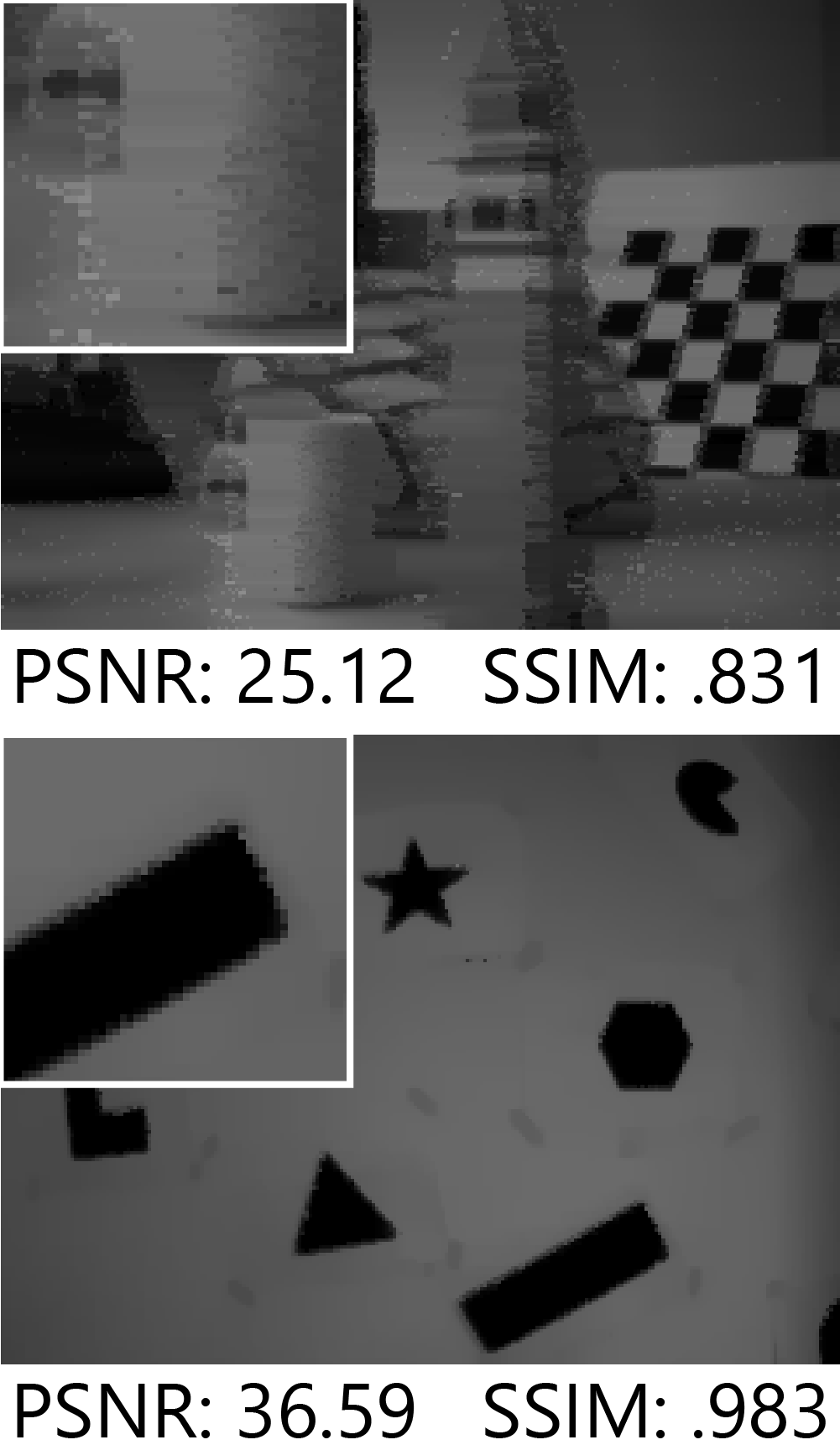}
        \caption{Ours (DMR)}
        \label{fig:pred_mbr_rec}
    \end{subfigure}%
    \begin{subfigure}[b]{0.16\linewidth}
        \includegraphics[width=.96\linewidth]{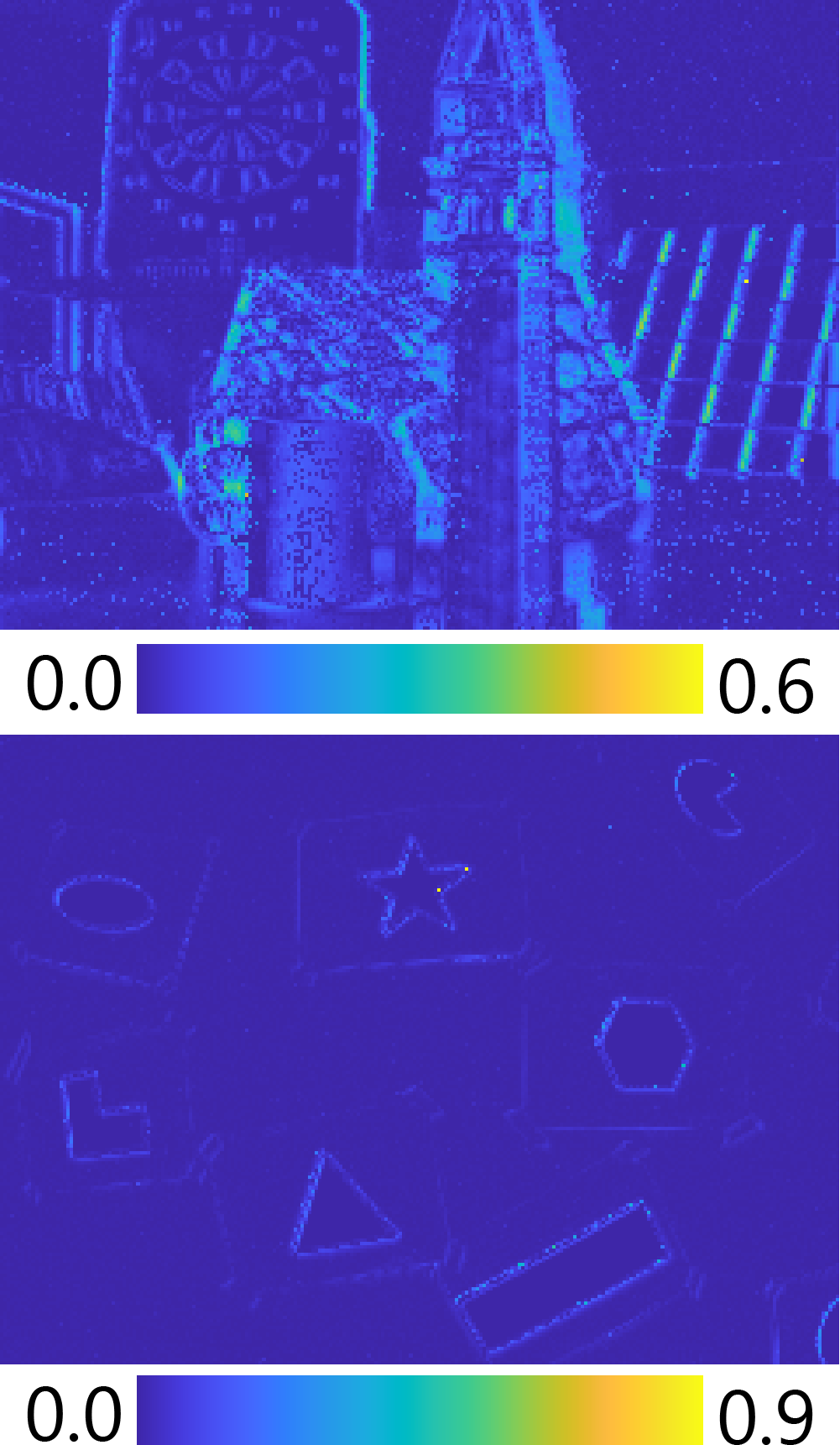}
        \caption{Error map of CF}
        \label{fig:pred_cf_err}
    \end{subfigure}%
    \begin{subfigure}[b]{0.16\linewidth}
        \includegraphics[width=.96\linewidth]{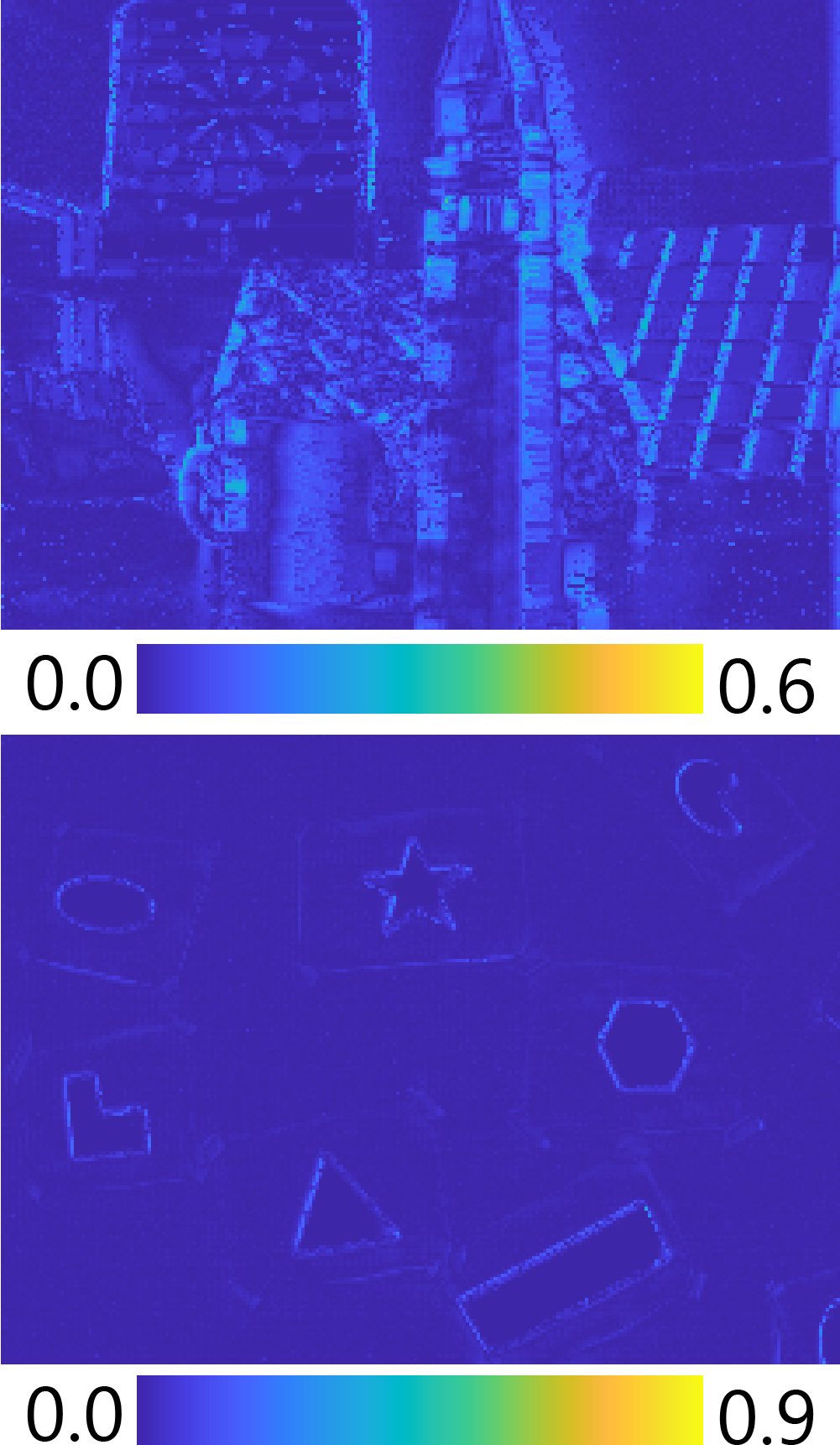}
        \caption{Error map of ours}
        \label{fig:pred_mbr_err}
    \end{subfigure}%
    \caption{Frame prediction. Given a start frame (a) and the future events (b) happened after (a), we predict the end frame (ground truth omitted). Our results using DMR alone outperforms existing algorithm, Complementary Filters (CF) \cite{Scheerlinck18accv}.}
    \label{fig:pred}
\end{figure*}

\customparagraph{Prediction.}\quad We next show frame prediction results, corresponding to Case 2 in \fref{fig:compare_int}. We withhold the end frame of two consecutive frames and seek to predict it using the start frame and ``future'' events. The results are shown in \fref{fig:pred}. Compared to CF \cite{Scheerlinck18accv}, our results are less noisy and closer to the ground truth.

\begin{figure}
     \centering
    \begin{subfigure}[t]{0.49\textwidth}
        \includegraphics[width=0.49\textwidth]{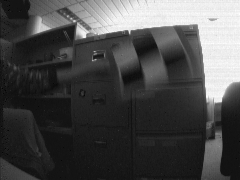}
        \includegraphics[width=0.49\textwidth]{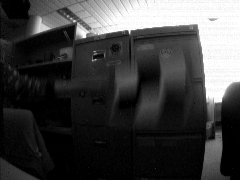}%
        \caption{Blurry images (data from \cite{Pan19cvpr})}
    \end{subfigure}
    \begin{subfigure}[t]{0.49\textwidth}
        \includegraphics[width=0.49\textwidth]{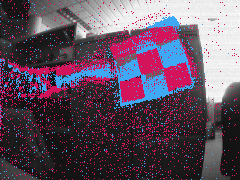}
        \includegraphics[width=0.49\textwidth]{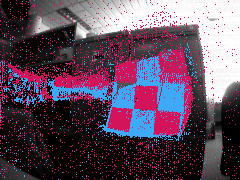}%
    \caption{w/ events during exposure time} 
    \end{subfigure}
    \begin{subfigure}[t]{0.49\textwidth}
        \includegraphics[width=0.49\textwidth]{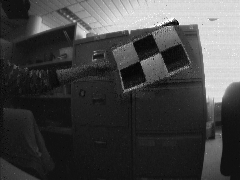}
        \includegraphics[width=0.49\textwidth]{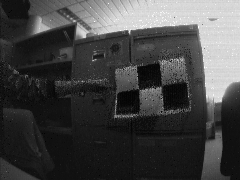}%
    \caption{EDI \cite{Pan19cvpr}} 
    \end{subfigure}
    \begin{subfigure}[t]{0.49\textwidth}
        \includegraphics[width=0.49\textwidth]{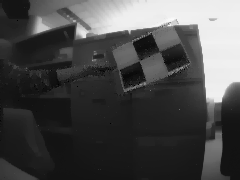}
        \includegraphics[width=0.49\textwidth]{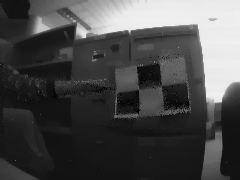}%
    \caption{Ours (DMR)} 
    \end{subfigure}
    \caption{Motion deblur. A motion blurred image (a) and the events during exposure time (b) are used to reconstruct a high framerate video. Compare to (c) EDI \cite{Pan19cvpr}, our results (d) preserves spatial features with less noise.}\label{fig:deblur}
\end{figure}

\customparagraph{Motion deblur.}\quad Corresponding to Case 3 in \fref{fig:compare_int}, we compare our DMR results with state-of-the-art, Event-based Double Integral (EDI) \cite{Pan19cvpr}, shown in \fref{fig:deblur}. Compared to EDI, our results preserves sharp edges while alleviating event noise.

\subsection{Results for RD}
\customparagraph{Data preparation.}\quad We use publicly available high-speed (240 FPS) video dataset, the Need for Speed dataset \cite{galoogahi2017need}. The reason we choose this dataset is because it has rich motion categories and content (100 videos with 380K frames) which involves both camera and scene/object motion. As introduced in \Sref{subsec:rd}, our RD is trained on the output of DMR process. As a proof of concept, we simulate solving a single-frame prediction problem, \ie given two consecutive video frames, we first simulate the latent event frame. Next, a DMR is performed to predict the end frame. 

\customparagraph{Training and testing.}\quad We randomly split the dataset into 89 training classes and 11 testing classes. For augmentation purpose, we perform a random temporal flip and a spatial crop with size $40\times40$. The sample clip will then experience event frame simulaltion and DMR using a random setting according to \Tref{tab:aug}. Note that we enforce generated event frames to contain less than 20\% of events. This is according to a statistical analysis of the DAVIS dataset\footnote{A statistical analysis is included in the supplementary material}. We generate 100K image pairs of size $40\times40$ pixels; 80\% of the sample dataset are randomly chosen as training samples and the rest 20\% are used for validation. We use a batch size of 128, which results in 2K batches per epoch. We use mini-batch stochastic gradient descent with an Adam optimizer ($\beta_1 = 0.9, \beta_2 = 0.999$). The learning rate is scheduled as $1\times10^{-3}$ for the initial 30 epochs, then $1\times10^{-4}$ for the following 30 epochs and $5\times10^{-5}$ afterwards. We use an NVIDIA TITAN X GPU for parallelization. Each epoch takes approximately 6 minutes training on our machine. We train our network for 150 epochs. Since our model is fully convolutional, the number of parameters is independent of the image size. This enables us to train on small patches ($40\times40$) and test on the whole image. 

\customparagraph{Plug \& play \emph{vs.} one-time denoising.}\quad Since we train our denoiser to establish a mapping function between DMR and its residual towards the ground truth, the first experiment we investigated is how/when to use this denoiser. We compare two frameworks, \ie, the plug \& play \cite{venkatakrishnan2013plug} and the one-time denoising. The plug \& play framework decouples the forward physical model and the denoising prior using the ADMM technique \cite{boyd2011distributed}. For one time denoising, we apply the residual denoiser once after the DMR has converged. One-time denoising is considered because it is considerably faster than plug \& play. Our experimental results show that one-time denoising performs similar or even better than plug \& play, shown in  \Tref{tab:pnp}. We reason that this is related to our training process and the initialization of the high-res tensor. Our differentiable model involves a temporal transition process from an existing frame to a future frame. We initialize the high-res tensor with the reference intensity frame. In each DMR iteration, the reconstruction process produces artifacts that are similar to the degradations in the initialized image. However, our denoiser is trained to ``recognize'' this degradation and remove these artifacts. Therefore, our denoiser is most useful and efficient when applied after the DMR has converged\footnote{Visual results are included in supplementary material.}. 

\begin{table}
    \caption{Plug \& play \emph{vs.} one-time denoising using RD.}\label{tab:pnp}
    \centering
    \begin{tabular}{| c | c | c |}
    \hline
    clip name & plug \&play & one-time \\
    \hline\hline
    Motorcycle & 28.07 / .951  & $\bm{29.11 / .965}$ \\
    Car race & 24.53 / .883 & $\bm{24.89 / .895}$ \\
    Football Player & 29.94 / .935 & $\bm{32.30 / .978}$ \\
    \hline
    \end{tabular}
    
\end{table}

\begin{figure*}
    \centering
    \begin{subfigure}[b]{0.2\linewidth}
        \includegraphics[width=.95\linewidth]{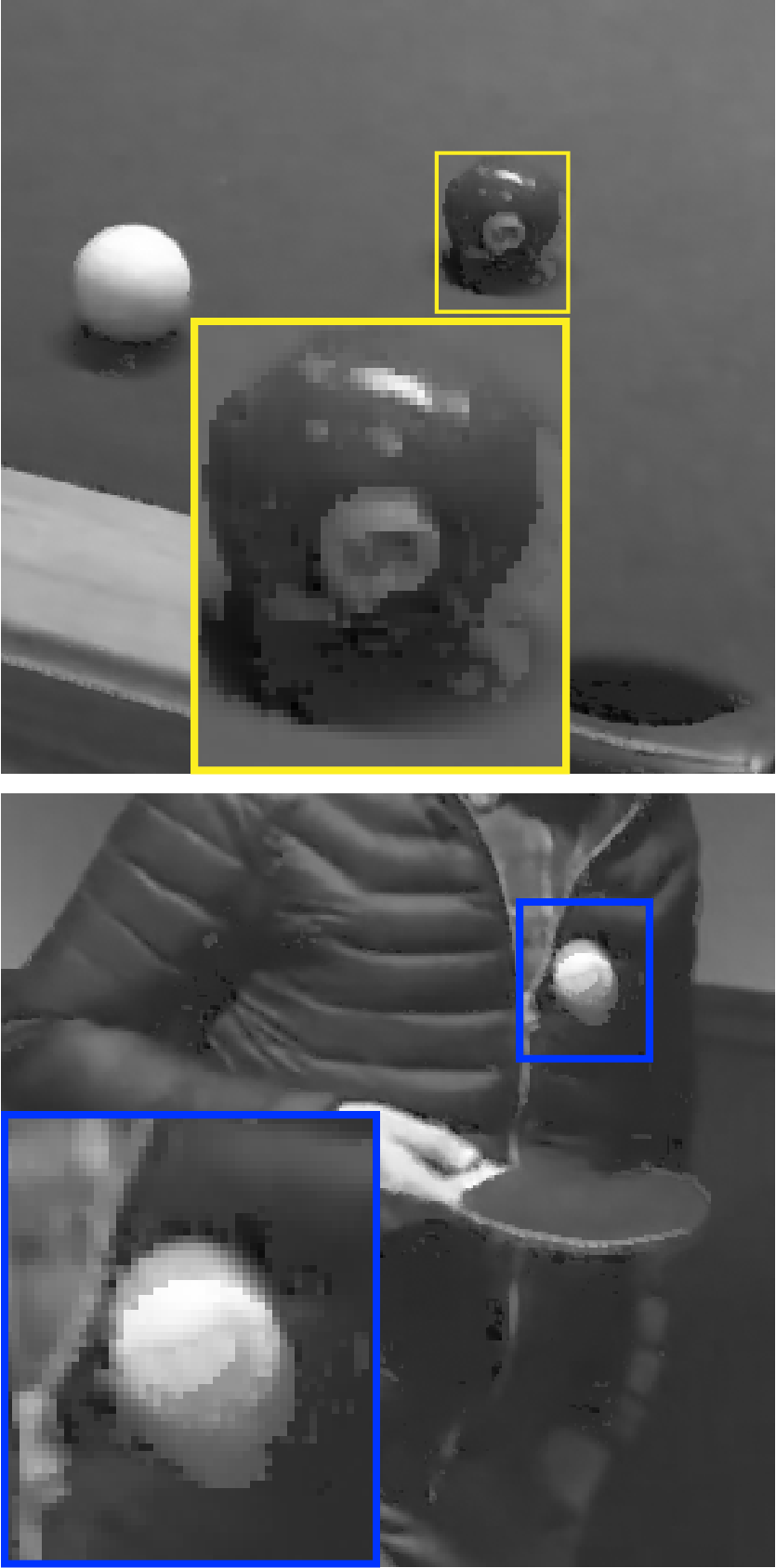}
        \caption{Ours (DMR)}
        \label{fig:dn_dmr}
    \end{subfigure}%
    \begin{subfigure}[b]{0.2\linewidth}
        \includegraphics[width=.95\linewidth]{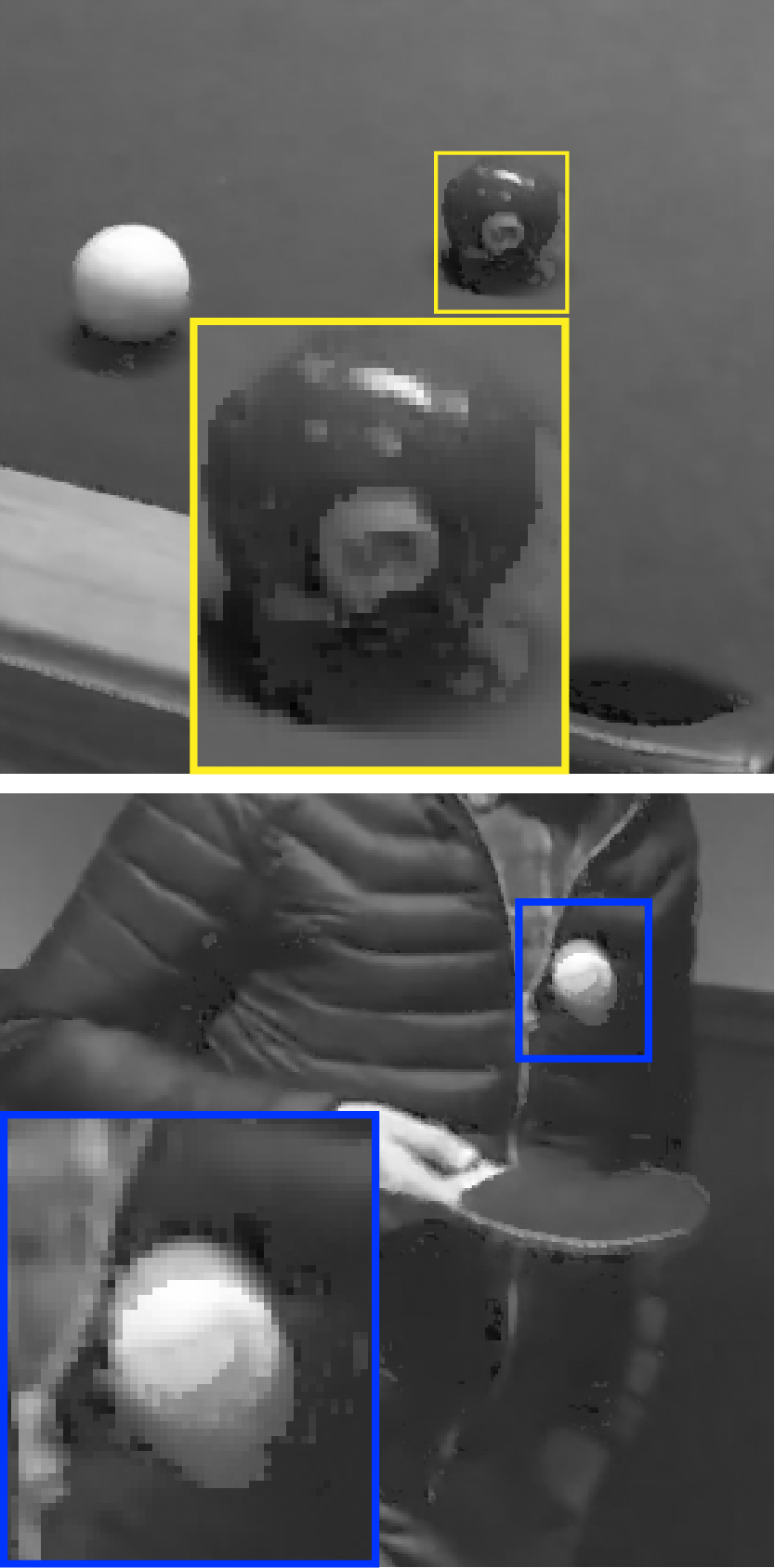}
        \caption{DnCNN \cite{zhang2017beyond}}
        \label{fig:dn_dncnn}
    \end{subfigure}%
    \begin{subfigure}[b]{0.2\linewidth}
        \includegraphics[width=.95\linewidth]{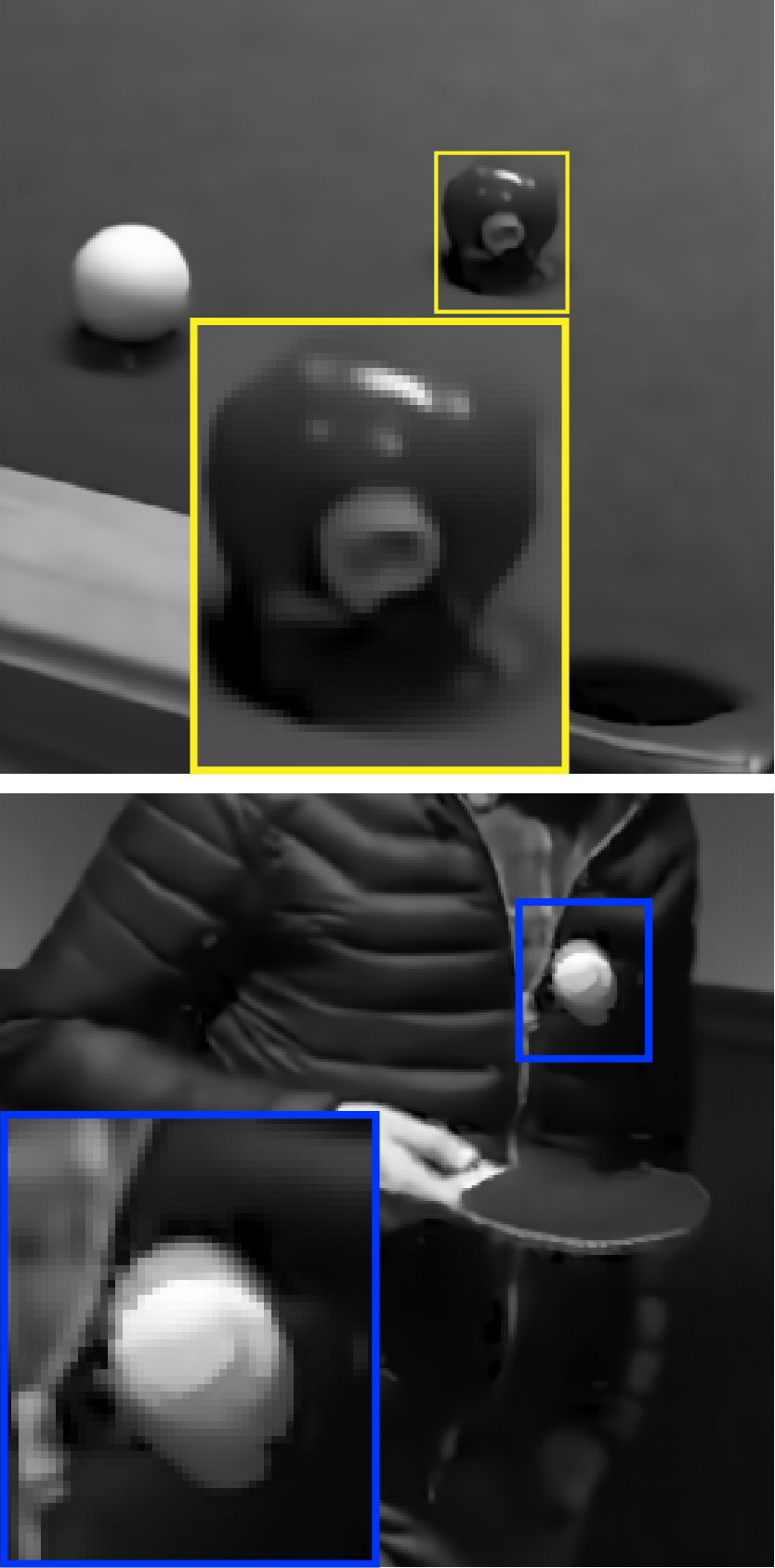}
        \caption{FFDNet \cite{zhang2018ffdnet}}
        \label{fig:dn_ffdnet}
    \end{subfigure}%
    \begin{subfigure}[b]{0.2\linewidth}
        \includegraphics[width=.95\linewidth]{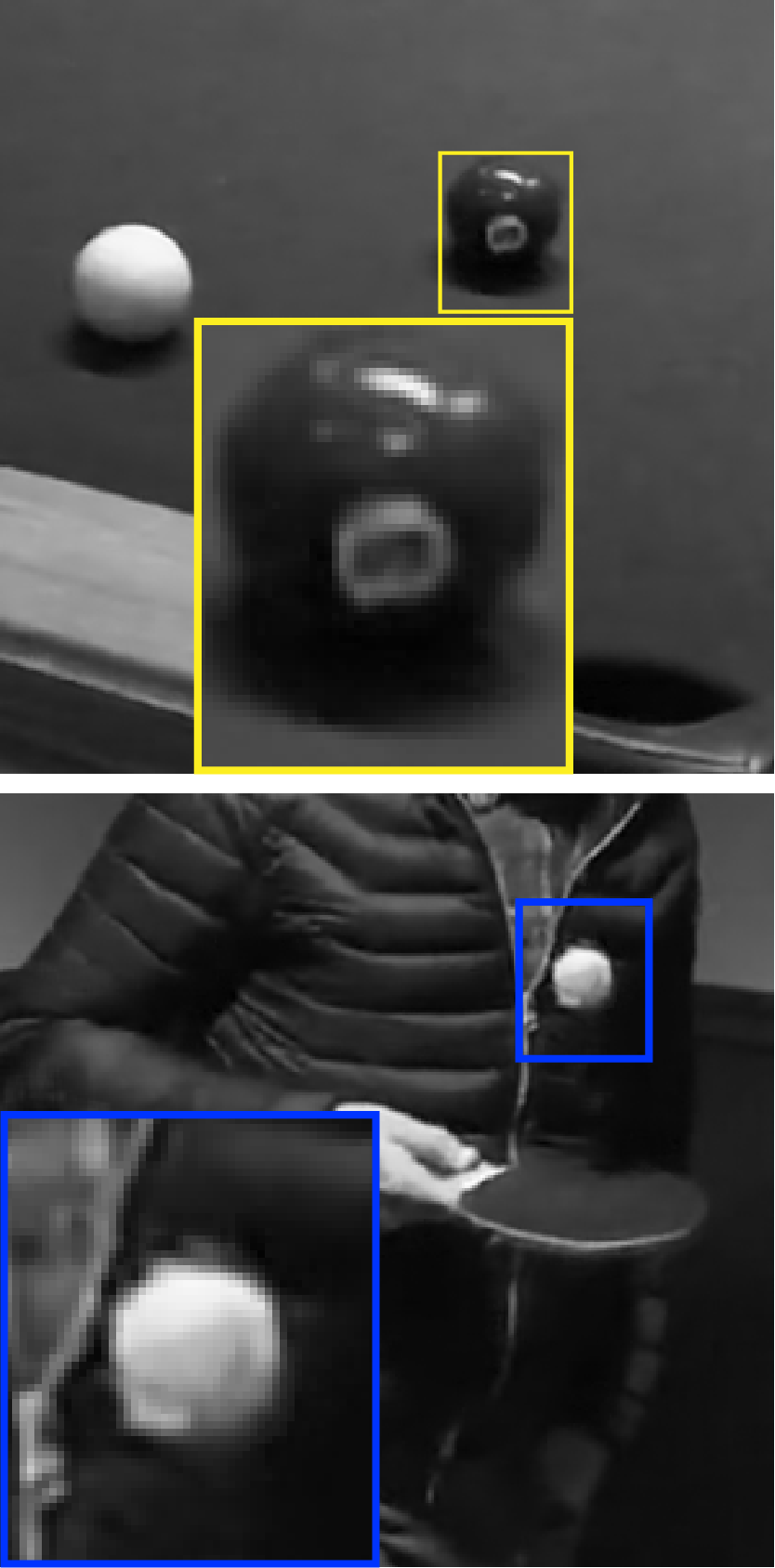}
        \caption{Ours (RD)}
        \label{fig:dn_ours}
    \end{subfigure}%
    \begin{subfigure}[b]{0.2\linewidth}
        \includegraphics[width=.95\linewidth]{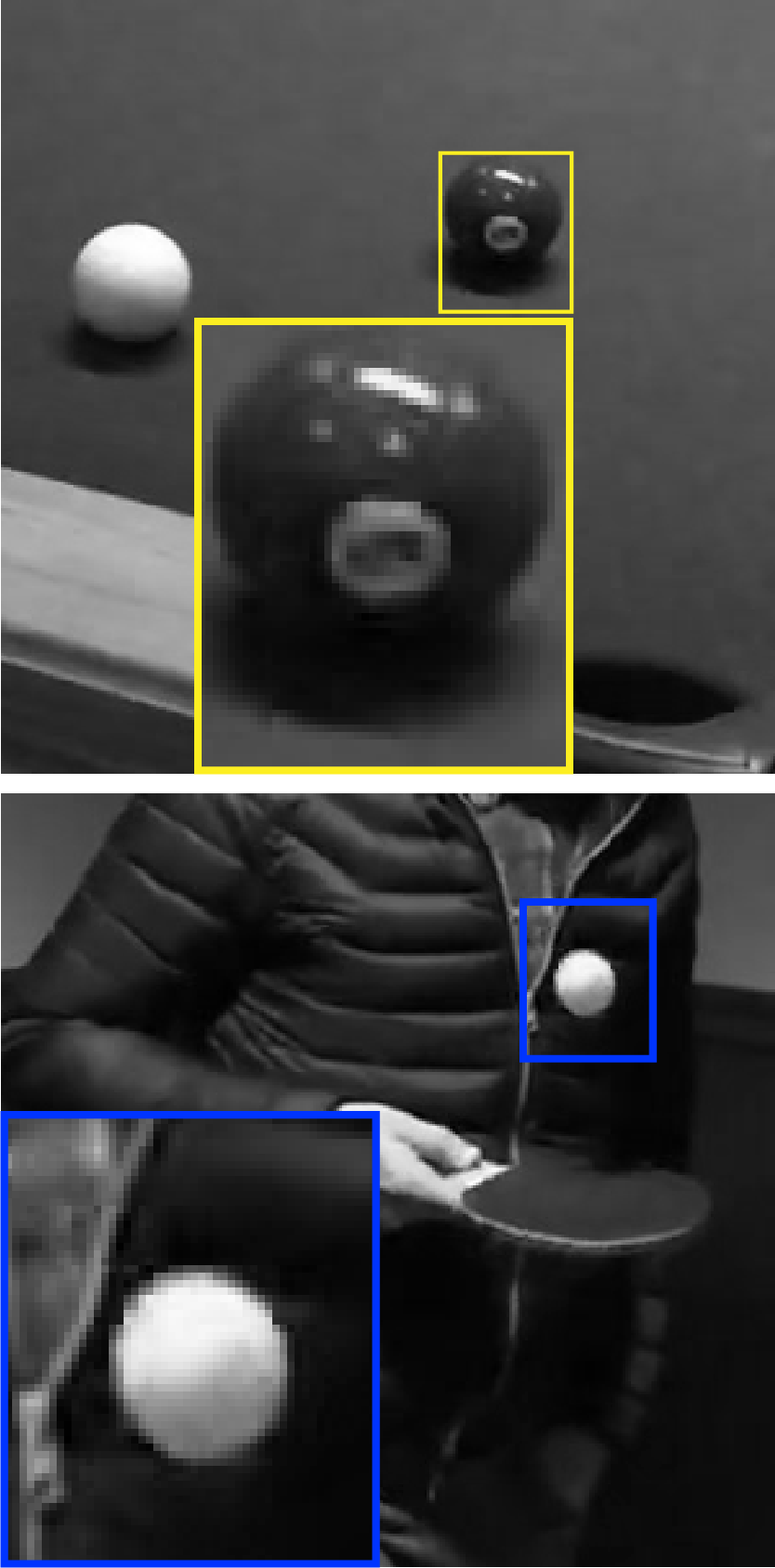}
        \caption{ground truth}
        \label{fig:dn_gt}
    \end{subfigure}%
    \caption{Comparison of denoising performance. Our learned Residual Denoiser (RD) reconstructs the intermediate frame (1-frame interpolation case) with fewer motion artifacts.}
    \label{fig:denoiser}
\end{figure*}

\customparagraph{Comparison with Gaussian denoisers.}\quad Since we decouple the problem as DMR and RD process, it is interesting to see whether a general denoiser can complete this task. We select several video clips from the testing classes and compare our results with two other denoisers, DnCNN \cite{zhang2017beyond} and FFDNet \cite{zhang2018ffdnet}. DnCNN is an end-to-end trainable deep CNN for image denoising with different Gaussian noise levels, \eg, [0, 55]. During our testing of DnCNN we found that the pre-trained weights do not perform well. We retrained the network using the Need for Speed dataset with Gaussian noise. The FFDNet is a later variant of DnCNN with the inclusion of pre- and post-processing. During our tuning of the FFDNet, we found that smaller noise levels (a tunable parameter for using the model) result in better denoising performance in terms of PSNR and SSIM metrics. For each testing image, we present the best tuned FFDNet result (noise level less than 10) and compare with our proposed denoiser. The results are summarized in \Tref{tab:denoiser}. Partial results\footnote{Full results can be seen in the supplementary material.} with zoom-in figures are presented in  \fref{fig:denoiser}.

\begin{table}
    \caption{Performance comparison for different denoisers.}
    \label{tab:denoiser}
\begin{center}
\centerline{
\resizebox{1\columnwidth}{!}{
    \begin{tabular}{|c|c|c|c|c|c|}
    \hline
    clip name & metric & DMR & DnCNN & FFDNet & Ours\\
    \hline\hline
    \multirow{2}{*}{airplane} & PSNR & 30.91 & 31.10 & 30.92 & $\bm{31.38}$\\
    & SSIM & 0.975 & 0.982 & 0.976 & $\bm{0.982}$\\
    \hline
    \multirow{2}{*}{basketball} & PSNR & 23.55 & 24.05 & 23.47 &$ \bm{24.06}$\\
    & SSIM & 0.963 & 0.971 & 0.964 & $\bm{0.972}$\\
    \hline
    \multirow{2}{*}{soccer} & PSNR & 29.96 & 31.08 & 30.13 & $\bm{31.29}$\\
    & SSIM & 0.961 & 0.974 & 0.962 & $\bm{0.975}$\\
    \hline
    \multirow{2}{*}{billiard} & PSNR & 36.46 & 35.42 & $\bm{36.48}$ & 36.46\\
    & SSIM & 0.982 & 0.986 & 0.983 & $\bm{0.987}$\\
    \hline
    \multirow{2}{*}{ping pong} & PSNR & 32.46 & 32.26 &$ \bm{32.50}$ & 32.24\\
    & SSIM & 0.974 & 0.978 & 0.975 & $\bm{0.979}$\\
    \hline
    \end{tabular}
}
}
\end{center}

\end{table}

\subsection{Comparison to non-event-based approach}
\begin{figure*}
    \centering
    \includegraphics[width=\linewidth]{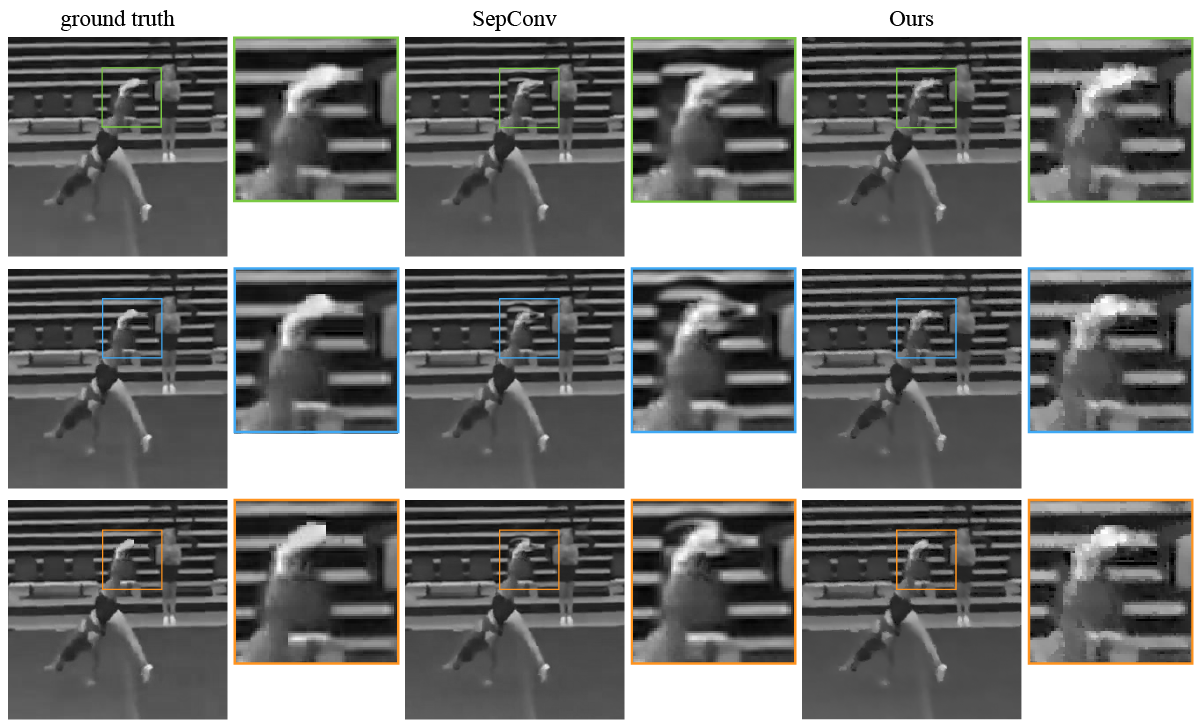}
    \caption{Multi-frame interpolation results, compared with SepConv \cite{niklaus2017video}. Shown are frames \#2, \#3 and \#4. Note that the intensity-only based frame interpolation method (SepConv) produces considerable motion artifacts around occluded areas, while our event-driven frame interpolation succesfully recovers image details in occluded regions.}
    \label{fig:interp}
\end{figure*}

We compared our results for performing multi-frame interpolation with a state-of-the-art approach, SepConv \cite{niklaus2017video}. We present results comparing 3-frame interpolation in \fref{fig:interp}. We convert our grayscale testing images to 3 channels (RGB) before applying the SepConv interpolation algorithm. Although the results from SepConv provide better visual experience, they have salient artifacts around large motion regions. Note that performing intensity only frame interpolation produces significant artifacts in the presence of severe occlusions. On the other hand, our event-driven frame interpolation is able to successfully recover image details in occluded regions of interpolated frames\footnote{Please see videos of results in supplementary material.}. For a quantitive comparison, the SepConv method has an average SSIM of 0.9566 and PSNR of 29.79. Ours have average SSIM of 0.9741 and PSNR of 37.64. 



\section{Concluding remarks}
In this paper, we have introduced a novel high framerate video synthesis framework by fusing intensity frames with event streams, taking advantages from both ends. Our framework includes two key steps, \ie, DMR and RD. Our DMR is free of training and is capable to unify different fusion settings between the two sensing modalities, which was not considered in previous work such as \cite{Pan19cvpr, Scheerlinck18accv}. We show in real data that our DMR performs better than existing algorithms. We show in simulation that a RD can be trained to effectively remove artifacts from DMR. Currently we train an RD from single-frame prediction case. It is interesting to further augment the training samples with \emph{all} the cases, which we will investigate in the future. Applying our RD to real data faces a domain gap due to the resolution (both spatial and temporal) and noise level mismatch. Currently, none of the existing DAVIS datasets contains enough sharp intensity images captured at high speed for training/fine-tuning. We will investigate event simulation using event simulator \cite{Rebecq18corl} in our future work.

{\small
\bibliographystyle{ieee}
\bibliography{ed-vfs-bib}
}

\end{document}